\documentclass{article}

\usepackage[latin1]{inputenc}
\usepackage{amsmath}
\usepackage{latexsym}
\usepackage{amssymb}
\usepackage{amsbsy}
\usepackage{rotating}

\def\CPP{\leavevmode\textrm{\hbox{C\hskip
-0.1ex\raise 0.5ex\hbox{\tiny ++}}}}

\newcommand{\csplib}[1]{%
(\texttt{prob{#1}} in~\cite{CSPLIB})}

\newcommand{\range}[2]{\left[#1\;..\;#2\right]}

\newcommand{\gfp}{\operatorname{gfp}}

\newcommand{\solv}{\operatorname{solv}}
\newcommand{\isolv}{\textsf{isolv}}
\newcommand{\prop}{\operatorname{prop}}
\newcommand{\iter}{\operatorname{iter}}
\newcommand{\dom}{\operatorname{dom}}

\newcommand{\new}{\mbox{\upshape\textsf{new}}}
\newcommand{\mychoose}{\mbox{\upshape\textsf{choose}}}

\newcommand{\init}{\operatorname{init}}

\newcommand{\ibnd}{\operatorname{zbnd}}
\newcommand{\rbnd}{\operatorname{rbnd}}
\newcommand{\vars}{\operatorname{vars}}
\newcommand{\ivars}{\operatorname{input}}
\newcommand{\ovars}{\operatorname{output}}
\newcommand{\idem}{\operatorname{idem}}
\newcommand{\sfix}{\operatorname{sfix}}
\newcommand{\dfix}{\operatorname{dfix}}
\newcommand{\setc}[2]{\{#1\;|\;#2\}}

\newcommand{\bigsqcap}{\mathop{\lower.1ex\hbox{\Large$\sqcap$}}}
\newcommand{\true}{\mathit{true}}
\newcommand{\false}{\mathit{false}}
\newcommand{\VV}{{\cal V}}
\newcommand{\RR}{{\mathbb R}}
\newcommand{\ZZ}{{\mathbb Z}}
\newcommand{\abounds}{\ensuremath{\operatorname{bounds}(\alpha)}}
\newcommand{\ibounds}{\ensuremath{\operatorname{bounds}(\ZZ)}}
\newcommand{\rbounds}{\ensuremath{\operatorname{bounds}(\RR)}}
\newcommand{\rangeop}{\operatorname{range}}

\newcommand{\bc}{\operatorname{bc}}
\newcommand{\lbc}{\operatorname{lbc}}
\newcommand{\ubc}{\operatorname{ubc}}
\newcommand{\fix}{\operatorname{fix}}
\newcommand{\dmc}{\operatorname{dmc}}
\newcommand{\two}{\operatorname{two}}
\newcommand{\ran}{\operatorname{ran}}
\newcommand{\neqe}{\operatorname{neq}}

\newcommand{\fran}{\operatorname{range}}

\newcommand{\events}{\operatorname{events}}
\newcommand{\devents}{\operatorname{devents}}
\newcommand{\mevents}{\operatorname{mevents}}
\newcommand{\es}{\operatorname{es}}
\newcommand{\evset}{\operatorname{evset}}

\newcommand{\PPN}[1]{\texttt{\lowercase{#1}}}

\newtheorem{theorem}{Theorem}[section]

\newtheorem{proposition}[theorem]{Proposition}
\newtheorem{ex}[theorem]{Example}
\newtheorem{dff}[theorem]{Definition}
\long\def\ignore#1{}
\newcommand{\pjs}[1]{\marginpar{\sc pjs}{\bf #1 }}
\newcommand{\schulte}[1]{\marginpar{\sc schulte}{\bf #1 }}

\newcommand{\BOX}[0]{$\Box$}

\long\def\short#1{}
        {\end{tabbing}\end{trivlist}}
\newenvironment{example}{\begin{ex}\rm}{\hfill\BOX\end{ex}}

\newenvironment{definition}{\begin{dff}\rm}{\end{dff}}

\newcommand{\calldiff}{\mbox{\upshape\ttfamily alldifferent}}
\newcommand{\cexactly}{\mbox{\upshape\ttfamily exactly}}
\newcommand{\cregular}{\mbox{\upshape\ttfamily regular}}
\newcommand{\clex}{\mbox{\upshape\ttfamily lex}}
\newcommand{\celement}{\mbox{\upshape\ttfamily element}}

\newcommand{\cA}{A}
\newcommand{\cB}{B}
\newcommand{\cC}{E}
\newcommand{\cG}{F}
\newcommand{\cH}{K}
\newcommand{\cI}{L}
\newcommand{\cJ}{O}
\newcommand{\cK}{P}
\newcommand{\cL}{N}
\newcommand{\cM}{Q}
\newcommand{\cN}{R}
\newcommand{\cO}{G}
\newcommand{\cP}{I}
\newcommand{\cQ}{J}
\newcommand{\cR}{C}
\newcommand{\cS}{D}
\newcommand{\cT}{M}
\newcommand{\cU}{N}
\newcommand{\cV}{H}

\newcommand{\citeyear}[1]{\cite{#1}}
\newenvironment{longitem}{\begin{itemize}}{\end{itemize}}
\newenvironment{proof}{\emph{Proof}:}{$\Box$}
\newenvironment{acks}{\subsection*{Acknowledgments}}{}

\begin{document}

\title{Efficient Constraint Propagation Engines}
\author{
CHRISTIAN SCHULTE\\
School of Information and Communication Technology, \\
KTH -- Royal Institute of Technology, \\
Isafjordsgatan 39, Electrum 229 \\
 SE-164 40 Kista, SWEDEN, \\ 
email: \texttt{cschulte@kth.se}.
\and
PETER J. STUCKEY\\
 NICTA Victoria Laboratory, Dept of Comp\@. Sci\@. \& Soft\@. Eng\@., \\
University of Melbourne 3010, AUSTRALIA,\\
 email: \texttt{pjs@cs.mu.oz.au}.
}
\date{}
\ignore{
[Constraints]
[Constraint and logic languages]


}

\maketitle

\begin{abstract}
  This paper presents a model and implementation techniques for
  speeding up constraint propagation.  Three fundamental
  approaches to improving constraint propagation based on
  propagators as implementations of constraints are explored:
  keeping track of which propagators are at fixpoint, choosing
  which propagator to apply next, and how to combine several
  propagators for the same constraint.
  
  We show how idempotence reasoning and events help track
  fixpoints more accurately.  We improve these methods by using
  them dynamically (taking into account current domains to
  improve accuracy).  We define priority-based approaches to
  choosing a next propagator and show that dynamic priorities can
  improve propagation.  We illustrate that the use of multiple
  propagators for the same constraint can be advantageous with
  priorities, and introduce staged propagators that combine the
  effects of multiple propagators with priorities for greater
  efficiency.
\end{abstract}

\ignore{
\begin{bottomstuff}
Author's addresses: C. Schulte,
School of Information and Communication Technology, KTH -- Royal Institute of Technology, 
Isafjordsgatan 39, Electrum 229
 SE-164 40 Kista, SWEDEN, email: \texttt{cschulte@kth.se}.
P.J. Stuckey, NICTA Victoria Laboratory, Dept of Comp\@. Sci\@. \& Soft\@. Eng\@., 
University of Melbourne 3010, AUSTRALIA, email: \texttt{pjs@cs.mu.oz.au}.
\permission
\copyright\ 2002 ACM 0164-0925/99/0100-0111 \$00.75
\end{bottomstuff}
}

\section{Introduction}

We consider the problem of solving 
{\em Constraint Satisfaction Problems\/} (CSPs) 
defined in the sense of Mackworth~\citeyear{mackworth}, which can be stated 
briefly as follows:
\begin{quote}
  We are given a set of variables, a domain of possible values for each
  variable, and a set (read as conjunction) of constraints.  Each 
  constraint is a relation
  defined over a subset of the variables, limiting the combination of
  values that the variables in this subset can take.  The goal is to
  find a {\em consistent\/} assignment of values to the variables so
  that all the constraints are satisfied simultaneously.
\end{quote}
One widely-adopted approach to solving CSPs combines 
backtracking tree search with constraint propagation. 
This framework is realized in finite domain 
constraint programming systems, 
such as SICStus Prolog~\cite{sicstus}, ILOG Solver~\cite{ILOG}, and
Gecode~\cite{gecode} that 
have been successfully applied to many real-life industrial 
applications.

At the core of a finite domain constraint programming system is a
constraint propagation engine that repeatedly executes
propagators for the constraints of a problem.  Propagators
discover and remove values from the domains of variables that
can no longer take part in a solution of the constraints.

\begin{example}
Consider a simple CSP, with variables $x_1$, $x_2$, and $x_3$
whose domain of possible values are respectively $x_1 \in \{2,3,4\}$,
$x_2 \in \{0,1,2,3\}$, $x_3 \in \{-1,0,1,2\}$  
and the constraints are $x_3 = x_2$, $x_1 \leq x_2 + 1$,
 and $x_1 \neq 3$. 

A propagator for $x_3 = x_2$ can determine that $x_2 \neq 3$
in any solution of this constraint 
since $x_3$ cannot take the value~3.
Similarly $x_3 \neq -1$.
The propagator then reduces the 
domains of the variables to
$x_2\in \{0,1,2\}$ and $x_3 \in \{0,1,2\}$.
A propagator for $x_1 \leq x_2 + 1$ can determine that
$x_1 \neq 4$ since $x_2 \geq 3$, and $x_2 \neq 0$ since
$x_1 \leq 1$ so the domains are reduced to
$x_1\in \{2,3\}$ and $x_2 \in \{1,2\}$. 
The propagator for $x_1 \neq  3$ can remove the value 3
from the domain of $x_1$, leaving $x_1 \in \{2\}$ (or $x_1 = 2$).
If we now reconsider the propagator for  $x_3 = x_2$ we can reduce
domains to $x_2 \in \{1,2\}$ and $x_3 \in \{1,2\}$.
No propagator can remove any further values.

We have not solved the problem, since we do not know a value for
each variable.  So after propagation we apply search, usually by
splitting the domain of a variable into two disjoint subsets and
considering the resulting two subproblems.

Suppose we split the domain of $x_2$.  One subproblem has
$x_1 \in \{2\}$, $x_2 \in \{1\}$ and $x_3 \in \{1,2\}$.
Applying the propagator for $x_2 = x_3$ results in
$x_3 \in \{2\}$.  Since each variable now takes a fixed value
we can check that $x_1 = 2, x_2 = 1, x_3 = 1$ is a solution
to the CSP.
The other subproblem has $x_1 \in \{2\}$, $x_2 \in \{2\}$ 
and $x_3 \in \{1,2\}$, and leads to another solution.
\end{example}
 
As can be seen from the example finite domain
constraint programming interleaves propagation with search. 
 In this paper we investigate how to
make a propagation engine as efficient as possible.

There are two important decisions the engine must make: 
which propagators should execute, and in which
order they should execute.
In order to make constraint propagation efficient, it is clear
that the engine needs to take the following issues
into account: avoid unnecessary propagator execution, restrict
propagation to relevant variables, and
choose the cheapest possible method for propagation.  In this
paper we show how propagation can be speeded up if the
engine takes these issues into account.

The contributions of the paper are as follows:
\begin{longitem}
\item We give a formal definition of propagation systems
  including fixpoint and event-based optimizations used in
  current propagation systems.
\item We extend event-based propagation systems to use
  dynamically changing event sets.
\item We introduce multiple propagators and staged propagators
  for a single constraint for use with propagation queues with
  priority.
\item We give experimental results that clarify the impact of
  many choices in implementing propagation engines: including
  idempotence reasoning, static and dynamic events, basic queuing
  strategies, priority queues, and staged propagation.
\end{longitem}

\paragraph*{Plan of the paper} 

The next section introduces
propagation-based constraint solving, followed by a model for
constraint propagation systems in Section~\ref{sec:systems}.
Section~\ref{sec:fixpoint} presents how to optimize propagation
by taking idempotence into account, while Section~\ref{sec:event}
explores the use of event sets. 
Which propagator
should be executed next 
is discussed
in Section~\ref{sec:prio}, while combination strategies
of multiple propagators for the same constraint is discussed in
Section~\ref{sec:mult}. 
Experiments for each feature are included in the relevant section,
and a summary is given in Section~\ref{sec:summary}.
Section~\ref{sec:conclusion} concludes.

\section{Propagation-based Constraint Solving}
\label{sec:prop}

This section defines our terminology for the basic components of a
constraint propagation engine.  In this paper we
restrict ourselves to finite domain 
integer constraint solving.
Almost all the discussion applies to other forms of finite domain
constraint solving such as for sets and multisets.

\paragraph*{Domains}

A \emph{domain} $D$ is a complete mapping from a fixed
(finite) set of variables $\VV$ to finite sets of integers.  A
\emph{false domain} $D$ is a domain with $D(x) = \emptyset$ for
some $x \in \VV$. 
\ignore{ 
Let $D_\bot(x) = \emptyset$ for all $x \in
\VV$. 
}
A variable $x\in\VV$ is \emph{fixed} by a domain $D$, if
$|D(x)|=1$. The \emph{intersection} of domains $D_1$ and $D_2$, denoted
$D_1 \sqcap D_2$, is defined by the domain $D(x) = D_1(x) \cap
D_2(x)$ for all $x \in \VV$.  

A domain $D_1$ is \emph{stronger}
than a domain $D_2$, written $D_1 \sqsubseteq D_2$, if $D_1(x)
\subseteq D_2(x)$ for all $x \in \VV$.
A domain $D_1$ is stronger than (equal to) a domain $D_2$ w.r.t.
variables $V$, denoted $D_1 \sqsubseteq_V D_2$ (resp.\ $D_1 =_V
D_2$), if $D_1(x) \subseteq D_2(x)$ (resp.\ $D_1(x) = D_2(x)$)
for all $x \in V$. 

A range is a contiguous set of integers, we use 
\emph{range} notation $\range{l}{u}$
to denote the range $\{ d \in \ZZ ~|~ l \leq d \leq u \}$ 
when $l$ and $u$ are integers.
A domain is a \emph{range domain} if $D(x)$ is a range for all $x$.
Let $D' = \rangeop(D)$ be the smallest range domain containing $D$, 
that is, the unique domain $D'(x) = \range{\inf D(x)}{\sup D(x)}$ for all $x \in \VV$.


We shall be interested in the notion of an \emph{initial domain},
which we denote $D_{\init}$.  The initial domain gives the initial
values possible for each variable. It allows us to restrict
attention to domains $D$ such that $D \sqsubseteq D_{\init}$.

\paragraph*{Valuations and constraints}

An \emph{integer valuation}  $\theta$ is a mapping of 
variables to integer values, written 
$\{x_1 \mapsto d_1, \ldots, x_n \mapsto d_n\}$.
We extend the valuation $\theta$ 
to map expressions and constraints involving the
variables in the natural way.

Let $\vars$ be the function that returns the set of
variables appearing in a valuation.
We define a valuation 
$\theta$ to be an element of a
domain $D$, written $\theta \in D$, if $\theta(x_i) \in D(x_i)$ for all
$x_i \in \vars(\theta)$.

The
\emph{infimum} and \emph{supremum} of an expression $e$ with
respect to a domain $D$ are defined as $\inf_D e = \inf~ \{
\theta(e) | \theta \in D\}$ and $\sup_D e = \sup~ \{ \theta(e) |
\theta \in D\}$.

We can map a valuation $\theta$ to a domain $D_\theta$ as follows
$$
D_\theta(x) = \left\{ \begin{array}{ll}
  \{ \theta(x) \} & x \in \vars(\theta) \\
  D_{\init}(x) & \mbox{otherwise} 
                      \end{array}
\right.
$$

A \emph{constraint} $c$ over variables $x_1, \ldots, x_n$
is a set of valuations $\theta$ such that $\vars(\theta) = \{x_1,\ldots,x_n\}$.
We also define $\vars(c) = \{x_1, \ldots, x_n\}$.

\paragraph*{Propagators}

We will \emph{implement} a constraint $c$ by a set of
propagators $\prop(c)$ that map domains to domains.  
A \emph{propagator} $f$ is a monotonically decreasing 
function from domains to domains: $f(D) \sqsubseteq D$,
and $f(D_1) \sqsubseteq f(D_2)$ whenever $D_1 \sqsubseteq D_2$.
A propagator $f$ is \emph{correct} for a constraint $c$ iff
for all domains $D$
$$
\setc{\theta}{\theta\in D} \cap c 
= 
\setc{\theta}{\theta\in f(D)} \cap c 
$$
This is a very weak restriction, for example the
identity propagator is correct for all constraints $c$.

A set of propagators $F$ is \emph{checking} for a constraint $c$,
if for all valuations $\theta$ where $\vars(\theta) = \vars(c)$
the following holds: $f(D_\theta) = D_\theta$ for all $f \in F$, iff 
$\theta \in c$. 
That is, for any domain $D_\theta$ corresponding to a valuation on $\vars(c)$,
$f(D_\theta)$ is a fixpoint iff $\theta$ is a solution of $c$.
We assume that $\prop(c)$ is a set of propagators that is
correct and checking for $c$.

\ignore{
\schulte{This is not needed, please remove}
The \emph{variables of a propagator} $\vars(f)$ are the variables
that can affect or be affected by the propagator.
It is defined as a set of variables $\vars(f)$ where
$D =_{\vars(f)} D'$ implies $f(D) =_{\vars(f)} f(D')$.
}

The \emph{output} variables $\ovars(f)\subseteq\VV$ of a
propagator $f$ are the variables changed by the propagator:
$x\in\ovars(f)$ if there exists a domain $D$ such that $f(D)(x)
\neq D(x)$.
The \emph{input} variables $\ivars(f)\subseteq\VV$ of a
propagator $f$ is the smallest subset $V \subseteq \VV$ such that
for each domain $D$: $D =_V D'$ implies that $D' \sqcap f(D)
=_{\ovars(f)} f(D') \sqcap D$.  
Only the input variables are
useful in computing the application of the propagator to the
domain. 

\begin{example}[\textbf{Propagators, input, and output}]
For the constraint $c \equiv x_1 \leq x_2 + 1$ 
the function $f_{\cA}$ defined by 
$f_{\cA}(D)(x_1) = \{ d \in D(x_1) ~|~ 
d \leq \sup_D x_2 + 1\}$ and $f_{\cA}(D)(v) = D(v), v \neq x_1$ 
is a correct propagator for $c$.  
Its output variables are $\{x_1\}$
and its input variables are $\{x_2\}$.
Let $D_1(x_1) = \{1, 5, 8\}$ and $D_1(x_2) = \{1, 5\}$, then
$f(D_1) = D_2$ where
$D_2(x_1) = D_2(x_2) = \{1, 5\}$.

The propagator $f_{\cB}$ defined as
$f_{\cB}(D)(x_2) = \{ d \in D(x_2) ~|~ d \geq \inf_D x_1 - 1\}$
and
$f_{\cB}(D)(v) = D(v), v \neq x_2$ 
is another correct propagator for $c$. Its output variables
are $\{x_2\}$ and input variables $\{x_1\}$.

The set $\{f_{\cA}, f_{\cB}\}$ is checking for $c$. The domain
$D_{\theta_1}(x_1) = D_{\theta_1}(x_2) = \{2\}$ corresponding to the
solution $\theta_1 = \{x_1 \mapsto 2,  x_2 \mapsto 2\}$ 
of $c$ is a fixpoint of both propagators. The
non-solution domain $D_{\theta_2}(x_1) = \{2\}$, $D_{\theta_2}(x_2) = \{0\}$ 
corresponding to the valuation $\theta_2 =  \{x_1 \mapsto 2,  x_2 \mapsto 0\}$ 
is not a fixpoint (of either propagator).
\end{example}

A \emph{propagation solver} $\solv(F,D)$
for a set of propagators 
$F$  and an initial domain $D$
finds the greatest mutual fixpoint of all the propagators
$f \in F$.
In other words, $\solv(F,D)$ returns a
new domain defined by
$$
\solv(F,D) = \gfp (\lambda d.\iter(F,d)) (D)
\hspace*{1cm}
\iter(F,D) = \bigsqcap_{f \in F} f(D) 
$$
where $\gfp$ denotes the greatest fixpoint w.r.t $\sqsubseteq$
lifted to functions.

Note that by inverting the direction of $\sqsubseteq$ we could 
equally well phrase this as a least fix point (as in~\cite{apt}).
But the current presentation emphasizes the \emph{reduction}
of domains as computation progresses.

\paragraph*{Domain and bounds propagators}

A consistency notion $C$ gives a condition on domains with
respect to constraints.  A set of propagators $F$ maintains
\emph{$C$-consistency} for a constraint $c$, if $\solv(F,D)$ is
always $C$ consistent for $c$.  Many propagators in practice are
designed to maintain some form of consistency: usually domain or
bounds.  But note that many more do not.

The most successful consistency technique 
was \emph{arc consistency}~\cite{mackworth},
which ensured that for each binary constraint, every value in the domain
of the first variable, has a supporting value in the
domain of the second variable that satisfied the constraint.
Arc consistency can be naturally extended to
constraints of more than two variables. This
extension has been called \emph{generalized arc consistency}~\cite{gac4},
as well as \emph{domain consistency}~\cite{ccfd,ccfd2}
(which is the terminology we will use),
and \emph{hyper-arc consistency}~\cite{book}.
A domain $D$ is \emph{domain consistent} for a constraint $c$
if $D$ is the least domain containing all solutions $\theta \in D$ of $c$,
that is, there does not exist $D' \sqsubset D$ such that
$\theta \in D \wedge \theta \in c \rightarrow \theta \in D'$.

Define the \emph{domain propagator} $\dom(c)$,
for a constraint $c$ as 
$$
\begin{array}{rcl@{\quad}l}
\dom(c)(D)(x) & = & \{ \theta(x) ~|~ \theta \in D \wedge
\theta \in c \} &\mbox{where~} x \in \vars(c)\\
\dom(c)(D)(x) & = & D(x) &\mbox{otherwise}
\end{array}
$$

The basis of bounds consistency is to relax the consistency requirement
to apply only to the lower and upper bounds of the domain of
each variable $x$. 
There are a number of different notions of 
bounds consistency~\cite{bounds-corr},
we give the two most common here.

A domain $D$ is \emph{\ibounds{} consistent} for a constraint $c$
with $\vars(c) = \{x_1, \ldots, x_n\}$,
if for each variable $x_i, 1 \leq i \leq n$ and 
for each $d_i  \in \{\inf_D x_i, \sup_D x_i\}$ there exist
\textbf{integers} $\boldsymbol{d_j}$ with 
$\inf_D x_j \leq d_j \leq \sup_D x_j$,
$1 \leq  j \leq n, j \neq i$ such that
$\theta = \{x_1 \mapsto d_1, \ldots, x_n \mapsto d_n\}$ is an 
\textbf{integer solution} of $c$. 

A domain $D$ is \emph{\rbounds{} consistent} for a constraint $c$
with $\vars(c) = \{x_1, \ldots, x_n\}$,
if for each variable $x_i, 1 \leq i \leq n$ and 
for each $d_i  \in \{\inf_D x_i, \sup_D x_i\}$ there exist
\textbf{real numbers} $\boldsymbol{d_j}$ with 
${\inf_D x_j \leq d_j \leq \sup_D x_j}$, 
$1 \leq  j \leq n, j \neq i$ such that
$\theta = \{x_1 \mapsto d_1, \ldots, x_n \mapsto d_n\}$ 
is a \textbf{real solution} of $c$. 

A set of propagators $F$ maintains 
\emph{\abounds{} consistency} for a constraint
$c$, if for all domains $D$, $\solv(F,D)$  is
\abounds{} consistent for $c$.

We can define a \emph{\ibounds{} propagator}, $\ibnd(c)$ 
for a constraint $c$ as
follows: 
\begin{eqnarray*}
\ibnd(c)(D) & = & D \sqcap \rangeop(\dom(c)(\rangeop(D)))
\end{eqnarray*}
It is not straightforward to give a generic description of 
\rbounds{} propagators, $\rbnd(c)$, for a constraint $c$,
that just maintains \rbounds{} consistency.
Examples~\ref{ex:fix} and~\ref{ex:nidem} define three such propagators.

\section{Constraint Propagation Systems}
\label{sec:systems}

A constraint propagation system evaluates the function
$\solv(F,D)$ during backtracking search.  We assume an execution
model for solving a constraint problem with a set of constraints
$C$ and an initial domain $D_0$ as follows.  We execute the
procedure $\textsf{search}(\emptyset,F,D_0)$ given in
Figure~\ref{fig:search} for an initial set
of propagators $F = \cup_{c \in C} \prop(c)$. This procedure is
used to make precise the optimizations presented in the
remainder of the paper.

\begin{figure}
\begin{tabbing}
xx\=xx\=xx\=xx\=xx\=choose $\{c_1, \ldots, c_m\}$  where 
              $C \wedge D \models c_1 \vee \cdots \vee c_m$\=\kill
\>\textsf{search}($F_o,F_n,D$) \\
\>\> $D$ := $\isolv(F_o,F_n,D)$ \> \> \>\> \% propagation \\
\>\> \textbf{if} ($D$ is a false domain)\\
\>\> \> \textbf{return} $\false$ \\
\>\> \textbf{if} ($\exists x \in \VV. |D(v)| > 1$) \\
\>\> \> choose $\{c_1, \ldots, c_m\}$  where 
              $C \wedge D \models c_1 \vee \cdots \vee c_m$ \>\>\> \% search strategy\\
\>\> \> \textbf{for} $i \in \range{1}{m}$ \\
\>\> \> \> \textbf{if} (\textsf{search}($F_o \cup F_n,\prop(c_i),D)$)\\
\>\> \> \> \> \textbf{return} $\true$ \\
\>\> \> \textbf{return} $\false$ \\
\>\> \textbf{return} $\true$
\end{tabbing}
\caption{Search procedure}\label{fig:search}
\end{figure}

Note that the propagators are partitioned into two sets, the old
propagators $F_o$ and the new propagators $F_n$. The
\emph{incremental} propagation solver $\isolv(F_o,F_n,D)$ (to be
presented later) takes advantage of the fact that $D$ is
guaranteed to be a fixpoint of the old propagators.

The somewhat unusual definition of search is quite general.  The
default search strategy for many problems is to choose a variable
$x$ such that $|D(x)| > 1$ and explore $x = \inf_D x$ or $x \geq
\inf_D x + 1$. This is commonly thought of as changing the domain
$D$ for $x$ to either $\{\inf_D x\}$ or $\{ d \in D(x)
~|~ d > \inf_D x\}$.  This framework allows more general
strategies, for example $x_1 \leq x_2$ or $x_1 > x_2$.

\begin{figure}
\begin{tabbing}
xx\=xx\=xx\= $\qquad Q$ := $Q \cup \new{}(f,F,D,D')$ \= \% add propagators
                  $f' \in F$  $\ldots$ \kill
\>\isolv($F_o,F_n,D$) \\
\>\> $F$ := $F_o \cup  F_n$; $Q$ := $F_n$ \\
\>\> \textbf{while} ($Q \neq \emptyset$) \\
\>\> \> $f$ := $\mychoose(Q)$ \>\%  select next propagator to apply \\
\>\> \> $Q$ := $Q - \{f\}$; $D'$ := $f(D)$ \\
\>\> \> $Q$ := $Q \cup \new{}(f,F,D,D')$ \> \% add propagators
                  $f' \in F$  $\ldots$ \\
\>\> \> $D$ := $D'$ \> \%                  $\ldots$ not necessarily at fixpoint at $D'$ \\
\>\> \textbf{return} $D$
\end{tabbing}
\caption{Incremental propagation solver.}\label{fig:prop}
\end{figure}

The basic incremental propagation solver algorithm is given in
Figure~\ref{fig:prop}.  The algorithm uses a queue $Q$ of propagators
to apply.  Initially, $Q$ contains the new propagators. Each
time the while loop is executed, a propagator $f$ is deleted from
the queue, $f$ is applied, and then all propagators that may no
longer be at a fixpoint at the new domain $D'$ are added to the
queue. An invariant of the algorithm is that at the while
statement $f(D) = D$ for all $f \in F - Q$.

The propagation solver \isolv~leaves two components
undefined: $\mychoose(Q)$ chooses the propagator $f \in Q$
to be applied next; $\new{}(f,F,D,D')$ determines the set
of propagators $f' \in F$ that are not guaranteed to be at their
fixpoint at the domain $D'$.  The remainder of the paper
investigates how to best implement these two components.

\subsection{Basic Variable Directed Propagation}

The core aim of the constraint propagation solver $\solv(F,D)$ is
to find a domain that is a mutual fixpoint of all $f \in
F$.  The incremental solver $\isolv(F_o,F_n,D)$ already takes
into account that initially $D$ is a fixpoint of propagators $f
\in F_o$.  The role of $\new{}$ is (generally) to return
\emph{as few} propagators $f \in F$ as possible.

A basic definition of $\new{}$ is as follows
$$
\new_{\ivars}(f,F,D,D')= 
\setc{f' \in F}{\ivars(f') \cap \setc{x\in\VV}{D(x)\neq
  D'(x)} \neq  \emptyset}
$$
Here all propagators $f'$ are added whose input variable domains
have changed. By the definition of input variables, if none
of them have changed for $f'$, then $f'(D') = D'$ since $f'(D) = D$
if $f' \in F - Q$.

\begin{proposition}\label{prop:input}
$\new_{\ivars}$ maintains the invariant $f'(D) = D$ for
all $f' \in F - Q$ at the start of the \textbf{while} loop.
\end{proposition}
\begin{proof}
Consider $f' \in F - Q$. 
Then $f'(D) = D$ and if  
$D =_{\ivars(f')} D'$ 
we have that 
$D' \sqcap f'(D) =_{\ovars(f')} f'(D') \sqcap D$.
Then
$$
\begin{array}{rlll}
D' & = & D' \sqcap D & \mbox{since $D' \sqsubseteq D$} \\
   & = & D' \sqcap f'(D) & \mbox{since $D = f'(D)$} \\
   & =_{\ovars(f')} & f'(D') \sqcap D & \mbox{by definition of $\ivars(f')$}
   \\
   & = & f'(D') & \mbox{since $f'(D') \sqsubseteq D' \sqsubseteq D$}
\end{array}
$$
Now $D' =_{\ovars(f')} f'(D')$ implies $D' = f'(D')$ by the definition
of $\ovars(f')$.
Hence each $f'$ in $F - Q$ is at fixpoint at the start of the loop. 
\end{proof}

The incremental propagation solver
\textsf{isolv} with this definition of \new{} 
(assuming $F_o = \emptyset$) is more or less equivalent to the
propagation algorithms in~\cite{BenhamouHeterogeneous}
and~\cite[page 267]{apt}.

\begin{example}[\textbf{Incremental propagation}]\label{ex:fix}
Consider the problem with constraints $c_{\cR} \equiv 
x_1 = 2 x_2$ and $c_{\cS} \equiv x_1 = 3 x_3$
represented by the \rbounds{} propagators
$$
\begin{array}{rcl@{\quad}l}
f_\cR(D)(x_1) & = & D(x_1) \cap \range{2 \inf_D x_2}{2 \sup_D x_2}, \\
f_\cR(D)(x_2) & = & D(x_2) \cap \range{\lceil \frac{1}{2} \inf_D x_1 \rceil}
{\lfloor \frac{1}{2} \sup_D x_1 \rfloor}, \\
f_\cR(D)(x) & = & D(x)& x \not\in \{x_1,x_2\}
\\
\\
f_\cS(D)(x_1) & = & D(x_1) \cap \range{3 \inf_D x_3}{3 \sup_D x_3}, \\
f_\cS(D)(x_3) & = & D(x_3) \cap \range{\lceil \frac{1}{3} \inf_D x_1 \rceil}
{\lfloor \frac{1}{3} \sup_D x_1 \rfloor}, \\
f_\cS(D)(x) & = & D(x)& x \not\in \{x_1,x_3\},
\end{array}
$$
with initial domains $D(x_1) = \range{0}{17}$,
$D(x_2) = \range{0}{9}$, and 
$D(x_3) = \range{0}{6}$.
Initially no constraint is at fixpoint, so $Q = \{f_\cR,f_\cS\}$.
$f_\cR$ is selected initially, and
we execute the bounds propagator determining
$D(x_1) = \range{0}{16}$, $D(x_2) = \range{0}{8}$.  
Since $x_1$ has changed, both $f_\cR$ and $f_\cS$ are added to the queue.
Then $f_\cS$ is executed setting
$D(x_1) = \range{0}{15}$, $D(x_3) = \range{0}{5}$.
Again both constraints are re-queued.
$f_\cR$ is executed changing the domains to 
$D(x_1) = \range{0}{14}$, $D(x_2) = \range{0}{7}$. 
Then $f_\cS$ changes the domains to 
$D(x_1) = \range{0}{12}$, $D(x_3) = \range{0}{4}$.
Since $x_1$ has changed we have $Q = \{f_\cR,f_\cS\}$.
Now $f_\cR$ is executed for no change, and $f_\cS$ is executed for
no change. 
We have reached a fixpoint $D(x_1) = \range{0}{12}$,
$D(x_2) = \range{0}{6}$, and $D(x_3) = \range{0}{4}$.
\end{example}

\section{Fixpoint Reasoning}
\label{sec:fixpoint}

The propagation engine computes a mutual fixpoint
of all the propagators.  Clearly, if we can determine
that some propagators are at fixpoint without executing them
we can limit the amount of work required by the engine.

\subsection{Static Fixpoint Reasoning}

A propagator $f$ is \emph{idempotent} if $f(D) = f(f(D))$ 
for all domains $D$.  That is, applying $f$ to any
domain $D$ yields a fixpoint of $f$.

\begin{example}[\textbf{Idempotent propagator}]
The propagator $f_{\cC}$ defined by
$$
\begin{array}{rcl@{\quad}l}
f_{\cC}(D)(x_1) &=& \setc{ d \in D(x_1)}{\frac{3}{2} d \in
  D(x_2)} \\
f_{\cC}(D)(x_2) &=& \setc{ d \in D(x_2)}{\frac{2}{3} d \in
  D(x_1)} \\
f_{\cC}(D)(x) &=& D(x) & x \not\in \{x_1,x_2\}
\end{array}
$$
is the domain propagator for the constraint $3x_1= 2x_2$.
The propagator $f_{\cC}$ is idempotent.
\end{example}

It is not difficult to see that each domain propagator $\dom(c)$
is idempotent.

\begin{proposition}
For all constraints $c$ and domains $D$
$$\dom(c)(D) = \dom(c)(\dom(c)(D))$$
\end{proposition}
\begin{proof}
Consider $\theta \in c$ where $\theta \in D$. Then by definition
$\theta(x) \in \dom(c)(D)$ for all $x \in \vars(c)$.
Hence $\theta \in \dom(c)(D)$. Since $\dom(c)(D) \sqsubseteq D$
we have $\theta \in c \wedge \theta \in D$ iff
$\theta \in c \wedge \theta \in \dom(c)(D)$.
Hence $\dom(c)(D) = \dom(c)(\dom(c)(D))$.
\end{proof}


\begin{example}[\textbf{Non-idempotent propagators}]\label{ex:nidem}
While many propagators are idempotent, some widely used
ones are \emph{not} idempotent.
Consider the constraint $3 x_1 = 2 x_2$ and
the propagator $f_{\cG}$:
$$
\begin{array}{rcl@{\quad}l}
f_{\cG}(D)(x_1) & = & D(x_1) \cap \range{\lceil \frac{2}{3} \inf_D x_2
\rceil}{\lfloor \frac{2}{3} \sup_D x_2 \rfloor} \\
f_{\cG}(D)(x_2) & = & D(x_2) \cap \range{\lceil \frac{3}{2} \inf_D x_1
\rceil}{\lfloor \frac{3}{2} \sup_D x_1 \rfloor} \\
f_{\cG}(D)(x) & = & D(x) & x \not\in \{x_1, x_2\}
\end{array}
$$

In almost all constraint programming systems
$\prop(3 x_1 = 2 x_2)$ is $\{f_{\cG}\}$ where
$f_{\cG}$ is the \rbounds{} propagator
for $3 x_1 = 2 x_2$.
Now $f_{\cG}$ is not idempotent. Consider $D(x_1) = \range{0}{3}$
and $D(x_2) = \range{0}{5}$.
Then $D' = f_{\cG}(D)$ is defined by
$D'(x_1) = \range{0}{3} \cap \range{0}{\lfloor 10/3\rfloor} = \range{0}{3}$
and
$D'(x_2) = \range{0}{5} \cap \range{0}{\lfloor 9/2\rfloor} = \range{0}{4}$.
Now $D'' = f_{\cG}(D')$ is defined by
$D''(x_1) = \range{0}{3} \cap \range{0}{\lfloor 8/3 \rfloor} = \range{0}{2}$
and
$D''(x_2) = \range{0}{4} \cap
\range{0}{\lfloor 9/2\rfloor} = \range{0}{4}$. Hence $f_{\cG}(f_{\cG}(D))=D''\neq D'=f_{\cG}(D)$.
\end{example}

We can always create an idempotent propagator $f'$ from a propagator
$f$ by defining $f'(D) = \solv(\{f\},D)$.  Indeed, in some 
implementations (for example~\cite{form-cons})
$\prop(3x_1=2x_2)$ is defined as the fixpoint of applying
$f_{\cG}$. 

Assume that $\idem(f)=\{f\}$ if $f$ is
an idempotent propagator and $\idem(f)=\emptyset$ otherwise.
The definition of $\new{}$ is improved by
taking idempotence into account
$$
\new_{\sfix}(f,F,D,D')= \new_{\ivars}(f,F,D,D')-\idem(f)
$$
An idempotent 
propagator is never put into the queue after application.

Note that without the idempotence optimization each
propagator $f$ that changes the domain is likely to be executed
again to check it is at fixpoint.  Almost all constraint
propagation solvers take into account static fixpoint reasoning (for
example ILOG Solver~\cite{ILOG}, Choco~\cite{choco}, 
SICStus~\cite{sicstus}, and Gecode~\cite{gecode}). 
Some systems even only allow idempotent
propagators (for example Mozart~\cite{mozart}).

\subsection{Dynamic Fixpoint Reasoning}

Even if a propagator is not idempotent we can often determine
that $f(D)$ is a fixpoint of $f$ for a specific domain $D$.
For simplicity we assume a function $\fix(f,D)$
that returns $\{f\}$ if it can show that $f(D)$ is a fixpoint for
$f$ and $\emptyset$ otherwise (of course without calculating $f(f(D))$, otherwise we gain
nothing). In practice this will be included in the implementation
of $f$.
$$
\new_{\dfix}(f,F,D,D')= 
\new_{\ivars}(f,F,D,D')
-\fix(f,D)
$$

\begin{example}[\textbf{Dynamic idempotence}]
  For bounds propagation for linear equations on range domains we
  are guaranteed that the propagator is at a fixpoint if there is
  no rounding required in determining new endpoints~\cite[Theorem
  8]{form-cons}.
  
  We can define $\fix(f_{\cG},D)$ for the bounds propagator
  $f_\cG$ from Example~\ref{ex:nidem} for the constraint $3x_1 =
  2x_2$, as returning $\{f_\cG\}$ if any new bound does not
  require rounding, e.g.~$2 \inf_D x_2/3 = \lceil 2 \inf_D x_2/3
  \rceil$ or $2 \inf_D x_2/3 \leq \inf_D x_1$ and similarly for
  the other three bounds.
  
  Consider applying $f_{\cG}$ to the domain $D''$ from the same
  example.  Now $D''' = f_{\cG}(D'')$ is defined by $D'''(x_1) =
  \range{0}{2} \cap \range{0}{\lfloor 8/3\rfloor} = \range{0}{2}$
  and $D'''(x_2) = \range{0}{4} \cap \range{0}{\lfloor
    6/2\rfloor} = \range{0}{3}$.  Notice that the new bound $x_2
  \leq 3$ is obtained without rounding $\lfloor 6/2\rfloor =
  \lfloor 3\rfloor = 3$. In this case we are guaranteed that the
  propagator is at a fixpoint.
\end{example}

Note that the dynamic case extends the static case since for
idempotent $f$ it holds that
$\fix(f,D) = \{f\}$  for all domains $D$.
The dynamic fixpoint reasoning extensions are obviously correct,
given  Proposition~\ref{prop:input}.

\begin{proposition}\label{prop:sidem}
$\new_{\dfix}$ maintains the invariant $f(D) = D$ for
all $f \in F - Q$ at the start of the \textbf{while} loop.
\end{proposition}

A complexity of either form of fixpoint reasoning is that a great
deal of care has to be taken when we claim a propagator is at
fixpoint, particularly for bounds propagators and for propagators
computing with multiple occurrences of the same variable.

\begin{example}[\textbf{Falling into domain holes}]
\label{ex:hole}
Consider a bounds propagator for $x_1 = x_2 + 1$ defined as
$$
\begin{array}{rcl@{\quad}l}
f_{\cO}(D)(x_1) &=& D(x_1) \cap \range{\inf_D x_2 + 1}{\sup_D x_2 + 1}\\
f_{\cO}(D)(x_2) &=& D(x_2) \cap \range{\inf_D x_1 - 1}{\sup_D x_1 - 1}\\
f_{\cO}(D)(x) &=& D(x) & x \not\in \{x_1,x_2\}
\end{array}
$$

We would expect this propagator to be idempotent since there is
no rounding required. Consider the application of $f_{\cO}$
to the domain $D(x_1) = \{0, 4,5,6\}$ and $D(x_2) = \{2,3,4,5\}$.
Then $f_{\cO}(D) = D'$ where 
$D'(x_1) = \{4,5,6\}$ and $D'(x_2) = \{2,3,4,5\}$.
This is not a fixpoint for $f_{\cO}$ because of the hole (that is
$1,2,4\not\in D(x_1)$) in the
original domain of $x_1$.
\end{example}

\begin{example}[\textbf{Multiple variable occurrences}]
\label{ex:alias} 
The $\cregular$ constraint introduced in~\cite{regular}
constrains a sequence of variables to take values described by a
regular expression (or a corresponding finite automaton). A
common case for the $\cregular$ constraint is to express cyclic
patterns by performing propagation on a sequence of variables
where some variables appear multiply.

Assume a propagator $f_{\cV}$ propagating that the sequence of
variables $\langle x_1,x_2,x_3\rangle$ conforms to the $\cregular$
expression $(11|00)0$ (that is, the values of three variables form either
the string $110$ or $000$). 

For a domain $D$ with $D(x_1)=D(x_2)=D(x_3)=\{0,1\}$ propagation
for the sequence $\langle x_1,x_2,x_3\rangle$ is obtained by
checking which values are still possible for each variable by
traversing the sequence once. In this particular case, for
$D'=f_{\cV}(D)$ we have that $D'(x_3)=\{0\}$ and
$D'(x_1)=D'(x_2)=\{0,1\}$ and $D'$ is a fixpoint for $f_{\cV}$.

Now assume a sequence $\langle x_1,x_2,x_1\rangle$ where $x_1$
appears twice. Using the same strategy as above, a single forward
traversal of the sequence yields: $D'=f_{\cV}(D)$ where
$D'(x_1)=\{0\}$ and $D'(x_2)=\{0,1\}$. However, $D'$ is
\emph{not} a fixpoint for $f_{\cV}$.
\end{example}

The two examples above describe common cases where a propagator
is not idempotent but in many cases still computes a fixpoint. In
these cases, dynamic fixpoint reasoning is beneficial: static
fixpoint reasoning would force these propagators to be never
considered to be at fixpoint while dynamic fixpoint reasoning
allows the propagator to decide whether the propagator is at
fixpoint for a given domain or not.

In practice dynamic fixpoint reasoning completely subsumes static
fixpoint reasoning and is easy to implement. To implement dynamic
fixpoint reasoning a propagator is extended to not only return
the new domain but also a flag indicating whether it is guaranteed
to be at fixpoint.

\subsection{Fixpoint Reasoning Experiments}

\begin{table}
\caption{Fixpoint reasoning experiments.}
\label{table:idem}

\begin{center}\footnotesize
\begin{tabular}{|l||rr|rr|rr|}
\hline
Example&\multicolumn{2}{c|}{none}&\multicolumn{2}{c|}{static}&\multicolumn{2}{c|}{dynamic}\\
&\multicolumn{1}{c}{time (ms)}&\multicolumn{1}{c|}{steps}&\multicolumn{1}{c}{time}&\multicolumn{1}{c|}{steps}&\multicolumn{1}{c}{time}&\multicolumn{1}{c|}{steps}\\\hline\hline
\texttt{all-interval-500}&$118.31$&$503\,036$&$-3.1\%$&$-24.8\%$&$-38.9\%$&$-24.9\%$\\\hline
\texttt{alpha}&$106.56$&$272\,398$&$-2.0\%$&$-5.7\%$&$-1.5\%$&$-5.7\%$\\\hline
\texttt{bibd-7-3-60}&$2\,279.36$&$1\,335\,657$&$-0.2\%$&$-6.5\%$&$-0.2\%$&$-6.5\%$\\\hline
\texttt{cars}&$4.64$&$15\,641$&$-0.4\%$&$\pm 0.0\%$&$+0.1\%$&$\pm 0.0\%$\\\hline
\texttt{crowded-chess-7}&$624.25$&$806\,664$&$-0.1\%$&$\pm 0.0\%$&$-0.2\%$&$\pm 0.0\%$\\\hline
\texttt{donald-b}&$0.70$&$546$&$-2.5\%$&$-8.4\%$&$-10.4\%$&$-16.5\%$\\\hline
\texttt{donald-d}&$30.37$&$46$&$-6.3\%$&$-6.5\%$&$-6.4\%$&$-13.0\%$\\\hline
\texttt{donald-v}&$0.38$&$546$&$-5.9\%$&$-16.5\%$&$-5.1\%$&$-16.5\%$\\\hline
\texttt{golomb-10-b}&$1\,347.48$&$2\,642\,464$&$+4.7\%$&$-17.9\%$&$+0.1\%$&$-17.5\%$\\\hline
\texttt{golomb-10-d}&$2\,430.00$&$2\,642\,962$&$+4.5\%$&$-18.0\%$&$+4.3\%$&$-18.0\%$\\\hline
\texttt{graph-color}&$35.87$&$9\,344$&$-0.1\%$&$\pm 0.0\%$&$-7.2\%$&$-6.2\%$\\\hline
\texttt{grocery}&$55.41$&$2\,299$&$+2.6\%$&$-3.8\%$&$+2.6\%$&$-3.8\%$\\\hline
\texttt{knights-10}&$7.46$&$48\,038$&$+0.9\%$&$+2.2\%$&$+0.9\%$&$+2.2\%$\\\hline
\texttt{minsort-200}&$342.48$&$240\,991$&$-11.3\%$&$-8.3\%$&$-5.1\%$&$-16.9\%$\\\hline
\texttt{o-latin-7-d}&$574.36$&$387\,060$&$\pm 0.0\%$&$-0.5\%$&$-8.7\%$&$-18.2\%$\\\hline
\texttt{partition-32}&$8\,571.24$&$16\,785\,128$&$-3.7\%$&$-18.8\%$&$-9.2\%$&$-19.7\%$\\\hline
\texttt{photo}&$108.87$&$422\,206$&$-0.3\%$&$-0.8\%$&$-14.6\%$&$-3.6\%$\\\hline
\texttt{picture}&$1\,553.42$&$150\,165$&$-4.1\%$&$-13.1\%$&$-3.4\%$&$-20.5\%$\\\hline
\texttt{queens-400}&$4\,433.12$&$31\,424\,152$&$+0.2\%$&$\pm 0.0\%$&$\pm 0.0\%$&$\pm 0.0\%$\\\hline
\texttt{queens-400-a}&$16.15$&$2\,469$&$-1.2\%$&$-16.2\%$&$-1.2\%$&$-16.2\%$\\\hline
\texttt{sequence-500}&$517.96$&$151\,609$&$-0.1\%$&$-0.1\%$&$-0.1\%$&$-0.1\%$\\\hline
\texttt{square-5-d}&$33\,391.24$&$1\,762\,492$&$-6.2\%$&$-16.0\%$&$-6.1\%$&$-16.1\%$\\\hline
\texttt{square-7-b}&$10\,166.24$&$7\,731\,557$&$+2.4\%$&$-14.5\%$&$-6.8\%$&$-14.7\%$\\\hline
\texttt{square-7-v}&$5\,690.00$&$13\,956\,982$&$-4.1\%$&$-16.9\%$&$-4.3\%$&$-16.9\%$\\\hline
\texttt{warehouse}&$0.74$&$2\,486$&$-2.5\%$&$-1.8\%$&$-7.9\%$&$-9.1\%$\\\hline
\hline
average (all)& --- & --- &$-1.6\%$&$-8.9\%$&$-5.6\%$&$-11.5\%$\\\hline
\end{tabular}
\end{center}

\end{table}

Table~\ref{table:idem} shows runtime (walltime) and the number of
propagation steps for the three different variants of 
fixpoint reasoning
considered. The first column presents absolute runtime values in
milliseconds 
and the number of propagation steps when no fixpoint
reasoning is
considered.  The two remaining columns ``static'' and ``dynamic''
show the relative change to runtime and propagation steps when
using static and dynamic fixpoint reasoning. 
For example, a relative
change of $+50\%$ means that the system takes $50\%$ more time or
propagation steps, whereas a value of $-50\%$ means that the
system takes only half the time or propagation steps. The row
``average (all)'' gives the average (geometric mean) of the relative
values for all examples.  More information on the examples
can be found in Appendix~\ref{sec:examples} and on the used
platform in Appendix~\ref{sec:platform}.

Note that both static and dynamic
fixpoint reasoning do not change the memory requirements for any of the
examples.

\paragraph*{Static fixpoint reasoning}

While static fixpoint reasoning reduces the number of propagator
executions by $8.9\%$ in average, the reduction in runtime is
modest by $1.6\%$ in average. The reason that the reduction in
propagation steps does not directly translate to a similar
reduction in runtime is that the avoided steps tend to
be cheap: after all, these are steps not performing any
propagation as the propagator is already at fixpoint. Examples
with significant reduction in 
runtime (such as \texttt{donald-d},
\texttt{minsort-200}, \texttt{picture}, and \texttt{square-5-d})
profit because the execution of costly propagators is avoided
(domain-consistent linear equations for \texttt{donald-d} and
\texttt{square-5-d}; $\cregular$ propagators for \texttt{picture};
minimum propagators involving up to 200 variables for
\texttt{minsort}). The behavior of \texttt{golomb-10-b} and
\texttt{golomb-10-d} is explained further below.

\paragraph*{Dynamic fixpoint reasoning}

As expected, both runtime as well as propagation steps are
considerably smaller for dynamic fixpoint reasoning compared to
static reasoning. This is in particular true for examples
\texttt{donald-b} and \texttt{square-7-b} where a
bounds-consistent \calldiff{} constraint can take advantage of
reporting whether propagation has computed a fixpoint due to no
domain holes as discussed in Example~\ref{ex:hole}
(\texttt{all-interval-500} shows the same behavior due to the
absolute value propagator used).


\paragraph*{Influence of propagation order}

Some examples show a considerable increase in runtime (in
particular, \texttt{golomb-10-b} and \texttt{golomb-10-d}). This
is due to the fact that the order in which propagators are
executed changes: costly propagators are executed often while
cheap propagators are executed less often (witnessed by the
decrease in propagation steps).

\begin{table}
\caption{Fixpoint reasoning experiments with propagator priorities.}
\label{table:idem-prio}

\begin{center}\footnotesize
\begin{tabular}{|l||rr|rr|}
\hline
Example&\multicolumn{2}{c|}{static}&\multicolumn{2}{c|}{dynamic}\\
&\multicolumn{1}{c}{time}&\multicolumn{1}{c|}{steps}&\multicolumn{1}{c}{time}&\multicolumn{1}{c|}{steps}\\\hline\hline
\texttt{all-interval-500}&$-2.9\%$&$-24.9\%$&$-37.7\%$&$-25.0\%$\\\hline
\texttt{alpha}&$-5.4\%$&$-15.6\%$&$-4.6\%$&$-15.6\%$\\\hline
\texttt{bibd-7-3-60}&$-0.7\%$&$-6.5\%$&$-0.7\%$&$-6.5\%$\\\hline
\texttt{cars}&$-0.2\%$&$\pm 0.0\%$&$+0.1\%$&$\pm 0.0\%$\\\hline
\texttt{crowded-chess-7}&$-0.7\%$&$\pm 0.0\%$&$-0.5\%$&$\pm 0.0\%$\\\hline
\texttt{donald-b}&$-4.8\%$&$-22.9\%$&$-11.8\%$&$-29.4\%$\\\hline
\texttt{donald-d}&$-6.7\%$&$-5.1\%$&$-6.7\%$&$-28.8\%$\\\hline
\texttt{donald-v}&$-5.6\%$&$-16.5\%$&$-5.3\%$&$-16.5\%$\\\hline
\texttt{golomb-10-b}&$-4.9\%$&$-23.5\%$&$-6.5\%$&$-23.5\%$\\\hline
\texttt{golomb-10-d}&$-3.3\%$&$-23.4\%$&$-3.0\%$&$-23.4\%$\\\hline
\texttt{graph-color}&$+0.4\%$&$\pm 0.0\%$&$-7.3\%$&$-4.9\%$\\\hline
\texttt{grocery}&$-4.9\%$&$-29.5\%$&$-4.9\%$&$-29.5\%$\\\hline
\texttt{knights-10}&$\pm 0.0\%$&$+1.3\%$&$+0.2\%$&$+1.3\%$\\\hline
\texttt{minsort-200}&$-10.9\%$&$-9.1\%$&$-0.9\%$&$-9.2\%$\\\hline
\texttt{o-latin-7-d}&$+0.3\%$&$-1.4\%$&$-6.1\%$&$-22.9\%$\\\hline
\texttt{partition-32}&$-8.5\%$&$-30.4\%$&$-11.8\%$&$-30.9\%$\\\hline
\texttt{photo}&$-0.7\%$&$-1.2\%$&$-7.5\%$&$-2.6\%$\\\hline
\texttt{picture}&$+6.5\%$&$-13.1\%$&$+4.0\%$&$-20.5\%$\\\hline
\texttt{queens-400}&$+0.2\%$&$\pm 0.0\%$&$+0.2\%$&$\pm 0.0\%$\\\hline
\texttt{queens-400-a}&$-1.3\%$&$-16.2\%$&$-1.3\%$&$-16.2\%$\\\hline
\texttt{sequence-500}&$+0.1\%$&$\pm 0.0\%$&$+0.1\%$&$\pm 0.0\%$\\\hline
\texttt{square-5-d}&$-6.4\%$&$-11.1\%$&$-7.3\%$&$-15.8\%$\\\hline
\texttt{square-7-b}&$-2.5\%$&$-25.2\%$&$-10.0\%$&$-26.0\%$\\\hline
\texttt{square-7-v}&$-6.6\%$&$-24.9\%$&$-5.8\%$&$-24.9\%$\\\hline
\texttt{warehouse}&$-2.2\%$&$-1.4\%$&$-7.9\%$&$-9.8\%$\\\hline
\hline
average (all)&$-2.9\%$&$-12.7\%$&$-6.1\%$&$-15.9\%$\\\hline
\end{tabular}
\end{center}

\end{table}

Section~\ref{sec:prio} presents priorities that order execution
according to propagator priorities. For now it is sufficient to
note that when these priority inversion problems are avoided
through the use of priorities, there is no increase in runtime.
Table~\ref{table:idem-prio} provides evidence for this. The table
shows relative runtime and propagation steps (relative to no
fixpoint reasoning as in Table~\ref{table:idem}). When avoiding
priority inversion it becomes clear that both static as well as
dynamic fixpoint reasoning consistently improve the number of
propagation steps and also runtime.  The only exception is
\texttt{picture} where even with priorities the considerable
reduction in execution steps does not translate into a reduction
in runtime. This is possibly due to a change in propagation order
that affects runtime (that this can have a remarkable effect even
with priorities is demonstrated in Section~\ref{sec:prio}).



\section{Event Reasoning}
\label{sec:event}

The next improvement for avoiding propagators to be put in the
queue is to consider what changes in domains of input variables
can cause the propagator to no longer be at a fixpoint. To this
end we use events: an \emph{event} is a change in the domain of a
variable.


Assume that the domain $D$ changes to the domain $D' \sqsubseteq
D$. A typical set of  events defined in a constraint propagation system
are:
\begin{itemize}
\item $\fix(x)$: the variable $x$ becomes fixed, that is
        $|D'(x)| = 1$ and $|D(x)| > 1$.
\item $\lbc(x)$: the lower bound of variable $x$ changes, that is
        $\inf_{D'} x > \inf_D x$.
\item $\ubc(x)$: the upper bound of variable $x$ changes,  that is
        $\sup_{D'} x < \sup_D x$.
\item $\dmc(x)$: the domain of variable $x$ changes, that is $D'(x) \subset
D(x)$.
\end{itemize}
Clearly the events overlap. Whenever a $\fix(x)$ event occurs
then a $\lbc(x)$ event, a $\ubc(x)$ event, or both events must also occur. 
If any of the first three events occur then a $\dmc(x)$ event occurs.
These events satisfy the following property.

\begin{definition}[\textbf{Event}]
An \emph{event} $\phi$ is a change in domain defined
by an event condition $\phi(D,D')$ 
which states that event $\phi$ occurs when the domain changes from
$D$ to $D' \sqsubseteq D$.
The event condition must satisfy the following 
property
$$
\phi(D,D'') = \phi(D,D') \vee \phi(D',D'')
$$
where $D'' \sqsubseteq D' \sqsubseteq D$.
So an event occurs on a change from $D$ to $D''$
iff it occurs in either the change from $D$ to $D'$ or
from $D'$ to $D''$.

Given a domain $D$ and a stronger domain $D' \sqsubseteq D$,
then $\events(D,D')$ is the set of events $\phi$
where $\phi(D,D')$. Suppose $D'' \sqsubseteq D' \sqsubseteq D$, then
clearly 
\begin{equation}\label{eq:events}
\events(D,D'') = \events(D,D') \cup \events(D',D'').
\end{equation}
\end{definition}

Most integer propagation solvers use the events defined above,
although many systems collapse $\ubc(x)$ and $\lbc(x)$ into a
single event $\bc(x)$ (for example, SICStus~\cite{sicstus},
ILOG Solver~\cite{ILOG}, and Gecode~\cite{gecode}). 
Choco~\cite{choco} maintains an event
queue and interleaves propagator execution with events causing
more propagators to be added to the queue.



Other kinds of events or variants of the above events are also
possible. For example, (for domains $D$ and $D'$ with $D'
\sqsubseteq D$):
\begin{itemize}
\item $\bc(x)$: as discussed above ($\lbc(x)\vee\ubc(x)$).
\item $\two(x)$: the variable $x$ reduces to a domain of at most
  two values: $|D'(x)| \leq 2$ and $|D(x)| > 2$.
\item $\ran(x)$: the variable $x$ reduces to a domain that will
  always be a range (two consecutive values or a single value):
  $\sup_{D'} x - \inf_{D'} x \leq 1$ and $\sup_{D} x - \inf_{D} >
  1$.
\item $\operatorname{pos}(x)$: the variable $x$ reduces to a
  domain that is strictly positive, that is $\inf_D' x > 0$ and
  $\inf_D x \leq 0$ (likewise, an event $\operatorname{neg}(x)$
  for reduction to a strictly negative domain).
\item $\operatorname{nneg}(x)$: the variable $x$ reduces to a
  domain that is non-negative: $\inf_D' x \geq 0$ and $\inf_D x
  < 0$ (likewise, an event $\operatorname{npos}(x)$ for reduction
  to a non-positive domain).
\item $\neqe(x,d)$: the variable $x$ can no longer take the value
  $d$, that is $d \in D(x)$ and $d \not\in D'(x)$
\end{itemize}

The events $\two$ and $\ran$ are useful for tracking
endpoint-relevance and
range-equivalence~\cite{SchulteStuckey:TOPLAS:2005}.  The $\neqe$
event has been used in e.g. Choco~\cite{choco} and
B-Prolog~\cite{Zhou:TPLP:05} for building AC4~\cite{ac4} style
propagators.

\begin{example}[\textbf{Events}]
Let  $D(x_1) = \{ 1, 2, 3 \}$, $D(x_2) = \{3,4,5,6\}$, 
$D(x_3) = \{ 0, 1\}$, and $D(x_4) = \{7,8,10\}$ while   
$D'(x_1) = \{ 1, 2 \}$, $D'(x_2) = \{3,5,6\}$,
$D'(x_3) = \{ 1\}$ and $D'(x_4) = \{7,8,10\}$. Then $\events(D,D')$ is
$$
\{ \ubc(x_1), \dmc(x_1), \dmc(x_2), \fix(x_3), \lbc(x_3), \dmc(x_3) \}
$$

Considering the additional events we obtain in addition
$$
\{ \bc(x_1), \two(x_1), \ran(x_1), \bc(x_3),
   \neqe(x_1,3), \neqe(x_2,4), \neqe(x_3,0) \}
$$
\end{example}


\begin{example}[\textbf{Events are monotonic}]
  Events are \emph{monotonic}: further changes to a domain do not
  discard events from previous changes. Consider the property
  $\fran(x)$ capturing that $D(x)$ is a range for a domain $D$
  (this property is related to the event $\ran(x)$, lacking the
  restriction that the domain can have at most two elements).
  
  The property $\fran(x)$ is \emph{not} an event: consider
  domains $D''\sqsubseteq D' \sqsubseteq D$ with
  $D(x)=\{1,2,3,5\}$, $D'(x)=\{1,2,3\}$, and $D''(x)=\{1,3\}$. If
  $\fran$ were an event, then
  $\events(D,D'')=\events(D,D')\cup\events(D',D'')$. However,
  $$
  \events(D,D'')=
  \{\dmc(x),\ubc(x)\}
  $$
  whereas
  $$
  \events(D,D')\cup\events(D',D'')=
  \{\dmc(x),\ubc(x),\fran(x)\}\cup\{\dmc(x)\}
  $$
\end{example}

\subsection{Static Event Sets}

Re-execution of
certain propagators can be avoided since they require certain
events to generate new information.

\begin{definition}[\textbf{Propagator dependence}]
A propagator $f$ is \emph{dependent on} a set of events $\es(f)$
iff
\begin{enumerate}
\item  for all domains $D$ if $f(D) \neq f(f(D))$ 
then $\events(D, f(D)) \cap \es(f) \neq \emptyset$,
\item for all domains $D$ and $D'$ where $f(D) = D$, 
$D' \sqsubseteq D$ and $f(D') \neq D'$
then $\events(D, D') \cap \es(f) \neq \emptyset$.
\end{enumerate}
\end{definition}

The definition captures the following. If $f$
is not at a fixpoint then one of the events in its event set
occurs. If $f$ is at a fixpoint $D$ then any change to a
domain that is not a fixpoint $D'$ involves an occurrence of one
of the events in its set.  Note that for idempotent propagators
the case~(a) never occurs.

For convenience later we will store the event set
chosen for a propagator $f$ in an array $\evset[f]$.

Clearly, if we keep track of the events since the last invocation
of a propagator, we do not need to apply a propagator if it is not
dependent on any of these events.

\begin{example}[\textbf{Event sets}]
Event sets for previously discussed propagators are as follows:
$$
\begin{array}{l@{\qquad}l}
f_{\cA} & \{\ubc(x_2)\} \\
f_{\cB} & \{\lbc(x_1)\} \\
f_{\cC} & \{\dmc(x_1), \dmc(x_2)\} \\
f_{\cG} & \{\lbc(x_1), \ubc(x_1), \lbc(x_2), \ubc(x_2)\} \\
\end{array}
$$
This is easy to see from the definitions of these
propagators. If they use $\inf_D x$ then $\lbc(x)$ is in
the event set, similarly if they use $\sup_D x$ then
$\ubc(x)$ is in the event set. If they use the entire domain $D(x)$
then $\dmc(x)$ is in the event set.
\end{example}

\emph{Indexical} propagation
solvers~\cite{ccfd2,CodognetDiaz:96,CarlssonOttossonEa:97} are based on such
reasoning. They define propagators in the form $f(D)(x) = D(x)
\cap e(D)$ where $e$ is an indexical expression.  The event set
for such propagators is automatically defined by the domain
access terms that occur in the expression $e$.

\begin{example}[\textbf{Indexical}]
An example of an indexical to propagate $x_1 \geq x_2 + 1$ is
$$
\begin{array}{rcl}
x_1 & \in & \range{\inf(x_2) + 1}{+\infty} \\
x_2 & \in & \range{-\infty}{\sup(x_1) - 1}
\end{array}
$$
These range expressions for indexicals define two propagators:
$$
\begin{array}{rcl@{\quad}l}
f_{\cP}(D)(x_1) &=& D(x_1) \cap \range{\inf(x_2) + 1}{+\infty}\\
f_{\cP}(D)(x) &=& D(x) & x \neq x_1\\
f_{\cQ}(D)(x_2) &=& D(x_2) \cap \range{-\infty}{\sup(x_1) - 1}\\
f_{\cQ}(D)(x) &=& D(x) & x \neq x_2
\end{array}
$$

The event set for the propagator $f_{\cP}$ from the definition
is $\{\lbc(x_2)\}$ while the the event set for
$f_{\cQ}$ is  $\{\ubc(x_1)\}$.
\end{example}

Using events we can define a much more accurate
version of \new{} that only adds propagators for
which one of the events in its event set has occurred.
$$
\new_{\events}(f,F,D,D') = 
      \setc{ f' \in F}{\evset[f'] \cap \events(D,D')\neq \emptyset } 
      - \fix(f,D)
$$
This version of \new{} (without dynamic fixpoint reasoning)
roughly corresponds with what most constraint propagation
systems currently implement.

\begin{proposition}\label{prop:events}
$\new_{\events}$ maintains the invariant $f(D) = D$ for
all $f \in F - Q$ at the start of the \textbf{while} loop.
\end{proposition}
\begin{proof}
Consider $f' \in F - Q - \{f\}$ different from the selected
propagator $f$.  Then $f'(D) = D$ and if $f'(D') \neq D'$
then $\events(D,D') \cap \es(f') \neq \emptyset$ by case (b)
of the definition
of $\es(f')$, so $f' \in Q$ at the start of the loop.

Consider selected propagator $f$. This is removed from $Q$,
but if $f(D') \neq D'$ then 
$\events(D,D') \cap \es(f) \neq \emptyset$ by case (a)
of the definition of $\es(f)$. 
Clearly also $\fix(f,D) \neq \emptyset$.
So $f \in Q$ at the start of the loop.
\end{proof}


\subsection{Event Set Experiments}
\label{sec:event-exp}

\begin{table}
\caption{Event set experiments (runtime and propagation steps).}
\label{table:static-event}

\begin{center}\footnotesize
\begin{tabular}{|l||rr|rr|rr|}
\hline
Example&\multicolumn{2}{c|}{only $\fix,\dmc$}&\multicolumn{2}{c|}{with $\bc$}&\multicolumn{2}{c|}{with $\lbc,\ubc$}\\
&\multicolumn{1}{c}{time}&\multicolumn{1}{c|}{steps}&\multicolumn{1}{c}{time}&\multicolumn{1}{c|}{steps}&\multicolumn{1}{c}{time}&\multicolumn{1}{c|}{steps}\\\hline\hline
\texttt{all-interval-500}&$+0.6\%$&$\pm 0.0\%$&$+1.0\%$&$\pm 0.0\%$&$+2.0\%$&$\pm 0.0\%$\\\hline
\texttt{alpha}&$-2.0\%$&$-4.4\%$&$-7.3\%$&$-19.2\%$&$-6.3\%$&$-19.3\%$\\\hline
\texttt{bibd-7-3-60}&$-2.6\%$&$-16.6\%$&$-1.4\%$&$-16.6\%$&$+7.4\%$&$-15.8\%$\\\hline
\texttt{cars}&$+1.1\%$&$-0.2\%$&$+1.7\%$&$-0.2\%$&$+2.6\%$&$-0.2\%$\\\hline
\texttt{crowded-chess-7}&$+8.3\%$&$\pm 0.0\%$&$+20.2\%$&$-8.4\%$&$+38.7\%$&$-8.4\%$\\\hline
\texttt{donald-b}&$+0.3\%$&$\pm 0.0\%$&$-2.1\%$&$-9.2\%$&$-1.3\%$&$-9.2\%$\\\hline
\texttt{donald-d}&$\pm 0.0\%$&$\pm 0.0\%$&$+0.9\%$&$\pm 0.0\%$&$-0.1\%$&$\pm 0.0\%$\\\hline
\texttt{donald-v}&$+0.6\%$&$-10.5\%$&$-1.4\%$&$-19.7\%$&$+0.1\%$&$-19.7\%$\\\hline
\texttt{golomb-10-b}&$-0.6\%$&$\pm 0.0\%$&$\pm 0.0\%$&$-3.9\%$&$-0.5\%$&$-3.9\%$\\\hline
\texttt{golomb-10-d}&$+0.2\%$&$\pm 0.0\%$&$+26.4\%$&$-3.5\%$&$+26.5\%$&$-3.5\%$\\\hline
\texttt{graph-color}&$-0.7\%$&$-47.7\%$&$+0.1\%$&$-49.5\%$&$+0.6\%$&$-49.5\%$\\\hline
\texttt{grocery}&$-0.1\%$&$\pm 0.0\%$&$\pm 0.0\%$&$\pm 0.0\%$&$\pm 0.0\%$&$\pm 0.0\%$\\\hline
\texttt{knights-10}&$-12.7\%$&$-47.5\%$&$-10.8\%$&$-48.6\%$&$-7.7\%$&$-48.6\%$\\\hline
\texttt{minsort-200}&$+0.4\%$&$\pm 0.0\%$&$+1.1\%$&$\pm 0.0\%$&$+1.5\%$&$\pm 0.0\%$\\\hline
\texttt{o-latin-7-d}&$-1.2\%$&$\pm 0.0\%$&$+0.2\%$&$-1.7\%$&$+1.5\%$&$-1.7\%$\\\hline
\texttt{partition-32}&$+0.7\%$&$\pm 0.0\%$&$+15.1\%$&$+3.4\%$&$+18.8\%$&$+17.7\%$\\\hline
\texttt{photo}&$+1.6\%$&$\pm 0.0\%$&$-3.3\%$&$-27.1\%$&$-2.0\%$&$-26.9\%$\\\hline
\texttt{picture}&$+2.0\%$&$\pm 0.0\%$&$+3.5\%$&$\pm 0.0\%$&$+11.5\%$&$\pm 0.0\%$\\\hline
\texttt{queens-400}&$-87.5\%$&$-99.1\%$&$-87.5\%$&$-99.1\%$&$-87.5\%$&$-99.1\%$\\\hline
\texttt{queens-400-a}&$-10.4\%$&$-38.8\%$&$-9.6\%$&$-38.8\%$&$-8.2\%$&$-38.8\%$\\\hline
\texttt{sequence-500}&$+2.2\%$&$\pm 0.0\%$&$+3.5\%$&$+36.3\%$&$+8.8\%$&$+36.3\%$\\\hline
\texttt{square-5-d}&$+0.8\%$&$\pm 0.0\%$&$+2.7\%$&$\pm 0.0\%$&$+2.6\%$&$+0.5\%$\\\hline
\texttt{square-7-b}&$-0.2\%$&$+0.5\%$&$+0.6\%$&$-8.2\%$&$+0.7\%$&$-7.8\%$\\\hline
\texttt{square-7-v}&$-1.0\%$&$-8.9\%$&$-1.7\%$&$-16.9\%$&$+1.2\%$&$-16.9\%$\\\hline
\texttt{warehouse}&$+2.2\%$&$+0.2\%$&$+3.3\%$&$-7.6\%$&$+4.2\%$&$-6.6\%$\\\hline
\hline
average (all)&$-8.5\%$&$-24.3\%$&$-6.7\%$&$-26.9\%$&$-4.6\%$&$-26.4\%$\\\hline
\end{tabular}
\end{center}

\end{table}

Table~\ref{table:static-event} shows runtime and number of
propagation steps for different event sets relative to a
propagation engine not using events (the engine uses dynamic
fixpoint reasoning but no priorities). The row ``average
(above)'' gives the geometric mean of the relative numbers given
in the table whereas ``average (all)'' shows the relative numbers
for all examples.

\paragraph*{General observations} 

A first, quite surprising, observation is that using no events at
all is not so bad.  It is the best approach for 10 out of the 25
benchmarks and for \texttt{crowded-chess-7} by a considerable
margin (between $8.3\%$ and $38.7\%$). Another general
observation is that a reduction in the number of propagation
steps does not directly translate into a reduction in runtime.
This is due to the fact that all saved propagator executions are
cheap: the propagator is already at fixpoint and does not have to
perform propagation. 

Note that the ratio between reduction in
steps and reduction in runtime ultimately depends on the
underlying system. In Gecode, the system used, the actual
overhead for executing a propagator is rather low. In systems
with higher overhead one can expect that the gain in runtime will
be more pronounced.

\paragraph*{Event set observations}

Adding the $\fix$ event is particularly beneficial for benchmarks
with many disequalities, particularly \texttt{queens-400} which only
uses disequalities.

Adding the $\bc$ event has significant benefit (up to 5\%) when
there are linear equalities as in \texttt{alpha} or \ibounds{} consistent
\calldiff{} as in \texttt{photo}. But for other examples, where one
would expect some reduction in runtime, the overhead (to be
discussed in more detail below) for maintaining a richer event
set exceeds the gains from reducing the number of propagation
steps. This is true for examples such as \texttt{minsort-200}
(minimum propagators), \texttt{partition-32} (multiplication), and
\texttt{golomb-10-b} and \texttt{square-7-b} (linear equations
and \ibounds{} consistent \calldiff).

Splitting the $\bc$ event into $\lbc$ and $\ubc$ events exposes the
overhead once more. There is almost never an improvement in
number of propagations, since only inequalities can actually
benefit, and there is substantial overhead.

\begin{table}
\caption{Event set experiments with priorities (runtime and
  propagation steps).}
\label{table:static-event-prio}
\begin{center}\footnotesize
\begin{tabular}{|l||rr|rr|rr|}
\hline
Example&\multicolumn{2}{c|}{only $\fix,\dmc$}&\multicolumn{2}{c|}{with $\bc$}&\multicolumn{2}{c|}{with $\lbc,\ubc$}\\
&\multicolumn{1}{c}{time}&\multicolumn{1}{c|}{steps}&\multicolumn{1}{c}{time}&\multicolumn{1}{c|}{steps}&\multicolumn{1}{c}{time}&\multicolumn{1}{c|}{steps}\\\hline\hline
\texttt{bibd-7-3-60}&$+6.1\%$&$-3.3\%$&$+0.6\%$&$-3.3\%$&$+13.1\%$&$-2.5\%$\\\hline
\texttt{crowded-chess-7}&$+8.0\%$&$\pm 0.0\%$&$+20.4\%$&$-0.7\%$&$+38.0\%$&$-0.7\%$\\\hline
\texttt{donald-b}&$+0.9\%$&$\pm 0.0\%$&$-1.1\%$&$-9.2\%$&$-0.7\%$&$-9.2\%$\\\hline
\texttt{donald-d}&$-0.3\%$&$\pm 0.0\%$&$+0.1\%$&$\pm 0.0\%$&$-0.2\%$&$\pm 0.0\%$\\\hline
\texttt{golomb-10-b}&$+0.1\%$&$\pm 0.0\%$&$-0.9\%$&$-5.5\%$&$-0.9\%$&$-5.5\%$\\\hline
\texttt{golomb-10-d}&$+0.1\%$&$\pm 0.0\%$&$-0.5\%$&$-7.8\%$&$\pm 0.0\%$&$-7.8\%$\\\hline
\texttt{minsort-200}&$+0.9\%$&$\pm 0.0\%$&$+1.5\%$&$\pm 0.0\%$&$+1.8\%$&$\pm 0.0\%$\\\hline
\texttt{partition-32}&$+1.9\%$&$-1.2\%$&$+1.8\%$&$-1.8\%$&$+1.0\%$&$-0.4\%$\\\hline
\texttt{picture}&$+2.6\%$&$\pm 0.0\%$&$+3.7\%$&$\pm 0.0\%$&$+1.6\%$&$\pm 0.0\%$\\\hline
\texttt{square-5-d}&$+0.9\%$&$\pm 0.0\%$&$+3.2\%$&$\pm 0.0\%$&$+3.2\%$&$+0.2\%$\\\hline
\texttt{square-7-b}&$+0.4\%$&$-0.1\%$&$+0.2\%$&$-10.6\%$&$+1.2\%$&$-10.6\%$\\\hline
\hline
average (above)&$+1.9\%$&$-0.4\%$&$+2.5\%$&$-3.6\%$&$+4.8\%$&$-3.4\%$\\\hline
average (all)&$-7.8\%$&$-24.1\%$&$-7.8\%$&$-27.8\%$&$-6.3\%$&$-27.7\%$\\\hline
\end{tabular}
\end{center}

\end{table}

\paragraph*{Influence of propagation order}

Similar to using fixpoint reasoning, the use of events also
changes the order in which propagators are executed.
Table~\ref{table:static-event-prio} reconsiders all examples that
could possibly benefit from $\bc$ or $\lbc,\ubc$ events and all
examples that show a remarkable increase in number of propagation
steps in Table~\ref{table:static-event}. 

The numbers confirm that with priorities no considerable increase
in runtime can be observed for all but \texttt{crowded-chess-7},
where the increase in runtime here is due to a change in
propagation order to the introduction of event sets that does not
depend on the relative priorities of the propagators used.

Once priorities are used, $\lbc$ and $\ubc$ are basically never
beneficial.

\paragraph*{Memory requirements}

\begin{table}
\caption{Event set experiments with priorities (allocated memory).}
\label{table:event-mem}
\begin{center}\footnotesize
\begin{tabular}{|l||r|r|r|r|}
\hline
Example&\multicolumn{1}{c|}{no events}&\multicolumn{1}{c|}{only $\fix,\dmc$}&\multicolumn{1}{c|}{with $\bc$}&\multicolumn{1}{c|}{with $\lbc,\ubc$}\\
&\multicolumn{1}{c|}{mem (KB)}&\multicolumn{1}{c|}{mem}&\multicolumn{1}{c|}{mem}&\multicolumn{1}{c|}{mem}\\\hline\hline
\texttt{bibd-7-3-60}&$6\,690.1$&$+2.9\%$&$+5.7\%$&$+14.4\%$\\\hline
\texttt{crowded-chess-7}&$198.2$&$+4.0\%$&$+13.1\%$&$+21.2\%$\\\hline
\texttt{donald-b}&$7.8$&$\pm 0.0\%$&$+12.8\%$&$+12.8\%$\\\hline
\texttt{donald-d}&$3.4$&$\pm 0.0\%$&$+58.2\%$&$+58.2\%$\\\hline
\texttt{donald-v}&$5.8$&$+34.3\%$&$+34.3\%$&$+51.5\%$\\\hline
\texttt{golomb-10-b}&$39.7$&$+2.6\%$&$+10.1\%$&$+12.7\%$\\\hline
\texttt{golomb-10-d}&$37.7$&$\pm 0.0\%$&$+2.8\%$&$+16.0\%$\\\hline
\texttt{minsort-200}&$32\,419.1$&$+1.1\%$&$+2.3\%$&$+17.4\%$\\\hline
\texttt{partition-32}&$160.3$&$+3.7\%$&$+9.4\%$&$+17.5\%$\\\hline
\texttt{picture}&$451.6$&$+17.7\%$&$+17.7\%$&$+30.1\%$\\\hline
\texttt{queens-400}&$30\,286.6$&$+0.2\%$&$+0.2\%$&$+0.2\%$\\\hline
\texttt{queens-400-a}&$567.0$&$+2.8\%$&$+4.2\%$&$+4.2\%$\\\hline
\hline
average (above)& --- &$+5.4\%$&$+13.3\%$&$+20.3\%$\\\hline
average (all)& --- &$+3.9\%$&$+9.9\%$&$+15.5\%$\\\hline
\end{tabular}
\end{center}

\end{table}

Table~\ref{table:event-mem} shows the total memory allocated for
different event sets relative to a propagation engine not using
events (the engine uses dynamic fixpoint reasoning and
priorities). It is important to note that the memory figures
reflect the amount of memory allocated which is bigger than the
amount of memory actually used. In particular, for small examples
such as \texttt{donald-*} the increase in allocated memory just
reflects the fact that an additional memory block gets
allocated.  

Using just a $\fix$ event increases the required memory by less
than $4\%$ in average, while using full event sets increases
the required memory by $15.5\%$ in average.

It must be noted that Gecode (the system used) has a particularly
efficient implementation of event sets: the implementation uses a
single pointer for an entry in an event set; entries for the same
variable and the same propagator but with different events still
use a single pointer. The memory overhead is due to the fact that
per variable and supported event, one single pointer for
bookkeeping is needed. Hence, the highest overhead can be
expected for examples with many variables but relatively few
propagators and small event sets (such as \texttt{bibd-7-3-60},
\texttt{crowded-chess-7}, and \texttt{picture}). The overhead
becomes less noticeable for examples with many propagators or
large event sets and few variables (such as \texttt{queens-400}
and \texttt{queens-400-a}).

\paragraph*{Summary}

In summary, while there is a compelling argument for $\fix$ events,
there is only a weak case for $\bc$ being supported, and $\lbc$ and
$\ubc$ should not be used.

\subsection{Dynamic Event Sets}

Events help to improve the efficiency of a
propagation-based solver. 
Just as we can improve the use of fixpoint reasoning by examining the
dynamic case, we can also consider dynamically updating
event sets as more information is known about the 
variables in the propagator.

\paragraph*{Monotonic event sets}

\begin{definition}[\textbf{Monotonic propagator dependence}]
\label{def:mono}
A propagator $f$ is \emph{monotonically dependent on} a set of events $\es(f,D)$ 
\emph{in the context of domain $D$} iff
\begin{enumerate}
\item
for all domains $D_0 \sqsubseteq D$ if $f(D_0) \neq f(f(D_0))$ 
then $\events(D_0, f(D_0)) \cap \es(f,D) \neq \emptyset$,
\item
for domains $D_0$ and $D_1$ where $D_0 \sqsubseteq D$, 
$f(D_0) = D_0$, 
$D_1 \sqsubseteq D_0$ and $f(D_1) \neq D_1$
then $\events(D_0, D_1) \cap \es(f,D) \neq \emptyset$.
\end{enumerate}
\end{definition}

Clearly given this definition $\es(f,D)$ is monotonically
decreasing with $D$.  The simplest kind of event reduction occurs
by subsumption.

\begin{definition}[\textbf{Subsumption}]
A propagator $f$ is \emph{subsumed} for domain $D$, if
for each domain $D' \sqsubseteq D$ we have $f(D') = D'$.
\end{definition}

A subsumed propagator makes no future contribution.  If $f$ is
subsumed by $D$ then $\es(f,D) = \emptyset$ and $f$ is never
re-applied.  Most current constraint propagation systems
take into account propagator subsumption.

\begin{example}[\textbf{Subsumption}]
  Consider the propagator $f_{\cA}$
and the domain  $D$ with $D(x_1) = \range{1}{3}$
  and $D(x_2) = \range{3}{7}$. Then the constraint holds for
all $D' \sqsubseteq D$ and $\es(f,D) = \emptyset$. 
\end{example}

Changing event sets can occur in cases other than
subsumption.

\begin{example}[\textbf{Minimum propagator}]
\label{ex:min}
Consider the propagator
$f_{\cH}$ for $x_0 = \min(x_1,x_2)$ defined by
$$
\begin{array}{rcl}
f_{\cH}(D)(x_0) & = & D(x_0) \cap 
     \range{\min(\inf_D x_1,\inf_D x_2)}{\min(\sup_D x_1,\sup_D x_2)} \\
f_{\cH}(D)(x_i) & = & D(x_i) \cap \range{\inf_D x_0}{+\infty} 
             \hfill i \in \{1,2\} \\
f_{\cH}(D)(x) & = & D(x) \hfill x \not\in \{x_0,x_1,x_2\}
\end{array}
$$
The static event set $\es(f_{\cH})$ is 
$\{\lbc(x_0), \lbc(x_1), \ubc(x_1), lbc(x_2), \ubc(x_2)\}$.
Note that this propagator is idempotent.

But given domain $D$ 
where $D(x_0) = \range{1}{3}$
and $D(x_2) = \range{5}{7}$ we know that modifying the
value of $x_2$ will never cause propagation.
A minimal definition of $\es(f_{\cH},D)$ is
$\{ \lbc(x_0), \lbc(x_1), \ubc(x_1) \}$.
\end{example}

\begin{example}[\textbf{\cexactly{} propagator}]
Another example is a propagator for the $\cexactly$
constraint~\cite{exactly}:
$\cexactly([x_1,\ldots,x_n],m,k)$ states that exactly $m$ out of the variables
$x_1,\ldots,x_n$ are equal to a value $k$.  
As soon as one of the
$x_i$ becomes different from $k$, all events for $x_i$ can be
ignored. 
Originally the events are $\dmc(x_i), 1 \leq i \leq m$, $\lbc(m)$, $\ubc(m)$
and $\dmc(k)$. 

Suppose a domain $D$ where
$D(k) = \{1,3,8\}$ and $D(x_3) = \{2,5,6,10,11,12\}$, then 
$x_3 \neq k$ and we know its contribution to the $\cexactly$
constraint. We can remove the event $\dmc(x_3)$ from the event set safely.
\end{example}

Other examples for monotonic event sets are propagators for the
$\clex$ constraint and the generalized $\celement$ constraint.
When using a variant of the $\clex$ propagator proposed
in~\cite{lex}, events can be removed as soon as the order among
pairs of variables being compared can be decided. For the
generalized $\celement$ constraint~\cite{element}
where the array elements are variables, events for a variable
from the array can be safely removed as soon as the variable
becomes known to be different from the result of the $\celement$
constraint.

Using monotonic dynamic event sets we can refine our 
definition of \new{} as follows.
\begin{quote}
\begin{tabbing}
xx \= xx \= xx \= xx \= \kill
$\new_{\mevents}$($f,F,D,D'$) \\
\> $F'$ := $\{ f' \in F ~|~ \evset[f'] \cap \events(D,D') \} - \fix(f,D)$ \\
\> $\evset[f]$ := $\es(f,D')$ \\
\> \textbf{return} $F'$
\end{tabbing}
\end{quote}
Every time a propagator $f$ is applied its event set is updated
to take into account newly available information.

A related idea is the ``type reduction'' of~\cite{typer} where
propagators are improved as more knowledge on domains (here
called types) becomes available. For example, the implementation
of $x_0 = x_1 \times x_2$ will be replaced by a more efficient
one, when all elements in $D(x_1)$ and $D(x_2)$ are non-negative.
Here we concentrate on how the event sets change.  The two ideas
could be merged as they are complementary.


\begin{proposition}\label{prop:mevents}
$\new_{\mevents}$ maintains the invariant $f(D) = D$ for
all $f \in F - Q$ at the start of the \textbf{while} loop.
\end{proposition}
\begin{proof}
The proof is almost identical to that for Proposition~\ref{prop:events}
since we are working
in a context if $\evset[f] = \es(f,D^*)$ then $D \sqsubseteq D^*$.

For propagators using the monotonic event sets, we have the 
invariant that $f \in F - Q$
iff $\evset[f] = \es(f,D^*)$ where $D^*$ is the result of
the last time we executed propagator $f$.

Suppose we have $f' \in F - Q - \{f\}$ 
where $f'(D') \neq D'$ and $f(D) = D$ then 
$\events(D^*,D') \cap \es(f,D^*) \neq \emptyset$ and
since $f' \not\in Q$ we have that 
$\events(D^*,D) \cap \es(f,D^*) = \emptyset$ otherwise we would
have placed $f$ in the queue already, hence
by the equation~(\ref{eq:events})
$\events(D,D') \cap \es(f,D^*) \neq \emptyset$ so $f' \in Q$
at the start of the loop.

Consider selected propagator $f$. Then it is removed from $Q$
but if $f(D') \neq  D'$ then clearly $\fix(f,D) = \emptyset$
and using case (a) of Definition~\ref{def:mono} 
we have that $\events(D,D') \cap \es(f,D) \neq \emptyset$.
Hence $f \in Q$ at the start of the loop.
\end{proof}

\paragraph*{Fully dynamic event sets}

Note that for many propagators 
we can be more aggressive in our definition of event sets
if we allow the event sets to change in a manner that is not necessarily
monotonically decreasing.

\begin{definition}[\textbf{General propagator dependence}]\label{def:idem}
A propagator $f$ is \emph{dependent on} a set of events 
$\es(f,D)$ 
in the context of domain $D$
if for all domains $D_1$ where $D_1 \sqsubset D$ and $f(D_1) \neq D_1$  
then $\events(D, D_1) \cap \es(f,D) \neq \emptyset$.
\end{definition}

Using fully dynamic event sets we can refine our 
definition of \new{} as follows.
\begin{quote}
\begin{tabbing}
xx \= xx \= xx \= xx \= \kill
$\new_{\devents}$($f,F,D,D'$) \\
\> $F'$ := $\{ f' \in F ~|~ \evset[f'] \cap \events(D,D') \} - \fix(f,D)$
\\
\> \textbf{if} ($\fix(f,D) = \emptyset$) \\
\> \> $F'$ := $F' \cup \{f\}$ \\
\> $\evset[f]$ := $\es(f,D')$ \\
\> \textbf{return} $F'$
\end{tabbing}
\end{quote}

Every time a propagator $f$ is applied its event set is updated
to take into account newly available information.
The only difficult case is that if the
definition of dynamic dependency does not capture the events
that occur when moving from $D$ to $D' = f(D)$.
If fixpoint reasoning cannot guarantee a fixpoint we need
to add $f$ to the queue.

Fully dynamic event sets are more powerful than 
monotonically decreasing event sets, but in general
they require reasoning about the new event sets 
each time the  propagator is run.

\begin{example}[\textbf{Fully dynamic events for minimum}]
Given the propagator $f_{\cH}$ from Example~\ref{ex:min}
and the domain $D$ where $D(x_0)  =  \range{0}{10}$,
$D(x_1) = \range{0}{15}$,
and $D(x_2) = \range{5}{10}$.
$D$ is a fixpoint of $f_{\cH}$
and a minimal set
$\es(f_{\cH},D)$ is $$\{ \lbc(x_0), \lbc(x_1), \ubc(x_1), \ubc(x_2) \}$$

While at $D' \sqsubseteq D$ where $D'(x_0) = \range{5}{9}$,
$D'(x_1) = \range{6}{9}$, and $D'(x_2) = \range{5}{10}$,
which is also a fixpoint, the minimal set
$\es(f_{\cH},D')$ is $$\{ \lbc(x_0), \ubc(x_1), \lbc(x_2), \ubc(x_2) \}$$

For the constraint $c_\cH$ we simply need to maintain a
$\lbc$ event for some variable $x_i$ in the right hand side with
the minimal $\lbc$ value.
\end{example}

\begin{proposition}\label{prop:devents}
$\new_{devents}$ maintains the invariant $f(D) = D$ for
all $f \in F - Q$ at the start of the \textbf{while} loop.
\end{proposition}
\begin{proof}
We have the 
invariant that $f \in F - Q$
iff $\evset[f] = \es(f,D^*)$ where $D^*$ is the result of
the last time we executed propagator $f$.

Suppose we have $f' \in F - Q - \{f\}$ 
where $f'(D') \neq D'$ and $f(D) = D$ then 
$\events(D^*,D') \cap \es(f,D^*) \neq \emptyset$ and
since $f' \not\in Q$ we have that 
$\events(D^*,D) \cap \es(f,D^*) = \emptyset$ otherwise we would
have placed $f'$ in the queue already, hence
by the equation~(\ref{eq:events})
$\events(D,D') \cap \es(f,D^*) \neq \emptyset$ so $f' \in Q$
at the start of the loop.

Consider the selected propagator $f$. The same reasoning cannot
apply since even though we know that 
$\events(D^*,D') \cap \es(f,D^*) \neq \emptyset$,
we have no guarantee that 
$\events(D,D') \cap \es(f,D^*)$ is not empty.
Now if $\fix(f,D) = \emptyset$ then possibly $f(D') \neq D'$,
but this will force $f \in Q$ by the start of the loop.
\end{proof}

Effectively the fully dynamic event sets approach relies
only on the dynamic fixpoint reasoning of the
propagator $f$ to handle what happens when moving
from $D$ to $D'$.

Fully dynamic event sets are closely related to the watched
literals approach to improving unit propagation in SAT
solving~\cite{chaff}.  Using watched literals, unit propagation
only considers a clause for propagation if one of two
\emph{watched} literals in the clause becomes false. Recently,
the idea of watched literals has been used in constraint
programming for the Minion solver~\cite{minion:wl}. Watched
literals differ from the events we concentrate on here since they
take into account values (similar to the $\neqe(x,a)$ event).
Note that
dynamic event sets do not usually have the property of watched
literals, in that they do not need to be updated on backtracking.

\subsection{Dynamic Event Sets Experiments}

\begin{table}
\caption{Dynamic event sets experiments (with priorities).}
\label{table:dyn-event}
\begin{center}\footnotesize
\begin{tabular}{|l||rrr|rrr|}
\hline
Example&\multicolumn{3}{c|}{monotonic}&\multicolumn{3}{c|}{fully dynamic}\\
&\multicolumn{1}{c}{time}&\multicolumn{1}{c}{steps}&\multicolumn{1}{c|}{mem}&\multicolumn{1}{c}{time}&\multicolumn{1}{c}{steps}&\multicolumn{1}{c|}{mem}\\\hline\hline
\texttt{bibd-7-3-60}&$+0.4\%$&$\pm 0.0\%$&$\pm 0.0\%$&$-5.0\%$&$-5.2\%$&$\pm 0.0\%$\\\hline
\texttt{crowded-chess-7}&$-24.0\%$&$\pm 0.0\%$&$-15.2\%$&$-28.2\%$&$-34.2\%$&$-17.8\%$\\\hline
\texttt{sequence-500}&$-82.3\%$&$-78.3\%$&$-49.5\%$&$-82.1\%$&$-78.3\%$&$-49.5\%$\\\hline
\texttt{o-latin-7-d}&$-0.7\%$&$\pm 0.0\%$&$\pm 0.0\%$&$-1.7\%$&$\pm 0.0\%$&$\pm 0.0\%$\\\hline
\hline
average (above)&$-39.5\%$&$-31.8\%$&$-19.1\%$&$-41.1\%$&$-39.3\%$&$-19.7\%$\\\hline
average (all)&$-10.8\%$&$-8.2\%$&$-4.1\%$&$-11.2\%$&$-9.9\%$&$-4.2\%$\\\hline
\end{tabular}
\end{center}

\end{table}

Table~\ref{table:dyn-event} shows the comparison of monotonic and
fully dynamic event sets to a propagation solver using static
event sets with $\{\fix,\bc,\dmc\}$ events and priorities to
avoid priority inversion as discussed before.  The table lists
only examples where dynamic event sets are used.

The propagators using monotonic event sets are as
follows: for \texttt{bibd-7-3-60}: $\clex$; for
\texttt{crowded-chess-7}: $\cexactly$ and $\celement$; for
\texttt{sequence-500}: $\cexactly$; for \texttt{o-latin-7-d}:
$\clex$. Clearly, monotonic event sets lead to a drastic
reduction in both runtime and memory usage, where it is worth
noting that the reduction in time is even more marked than the
reduction in propagation steps (each propagation step becomes
cheaper as smaller event sets must be maintained).

Fully dynamic event sets are considered in Boolean-sum
propagators used in \texttt{bibd-7-3-60} and
\texttt{crowded-chess-7}. The difference in improvement between
the two examples can be explained by the fact that
\texttt{crowded-chess-7} uses Boolean-sums as inequalities while
\texttt{bibd-7-3-60} uses Boolean-sums as equalities where
inequalities offer the potential for considerably smaller event
sets~\cite{minion:wl}.

Example~\texttt{bibd-7-3-60} provides another insight: the
drastic reduction in runtime observed in~\cite{minion:wl} by
using watched literals for Boolean-sum propagators in the Minion
solver is most likely not due to small event sets but to other
aspects. A possible aspect is the knowledge about which variables
have been modified when a propagator is executed.

\section{Which Propagator to Execute Next}
\label{sec:prio}

We now address how to define which propagator $f$ in the queue
$Q$ should execute first, that is how to define the
\mychoose{} function.

\ignore{
The simplest policies to implement are FIFO (First In First Out)
or LIFO (Last In First Out) 
queue of propagators. The only wrinkle from normal
queuing operations is that we dont want to reinsert
a propagator already in the queue.
\begin{quote}
\begin{tabbing}
xx \= xx \= xx \= xx \= \kill
Q \cup F \\
\> \textbf{while} $(f \in F)$ \\
\> \> $F$ := $F - \{f\}$ \\
\> \> \textbf{if} ($f \not\in Q$) $Q$ := \textsf{push}($Q$,$f$) \\
\> \textbf{return} $Q$ \\
\\
\textbf{choose}(Q)
\end{tabbing}
\end{quote}
The $push$ operator is either a stack or queue push.
}

The simplest policy to implement is a FIFO (First In First Out)
queue of propagators.
Propagators are added to the queue, if they are not already
present, and \mychoose{} 
selects the oldest propagator in
the queue.
The FIFO policy ensures fairness so that computation is not
dominated by a single group of propagators, while possibly not
discovering failure (a false domain) from other propagators
quickly.

The equally simple LIFO (Last In First Out) policy is a stack
where propagators not already in the stack are pushed, and
\mychoose{} selects the top of the stack.

\ignore{
\pjs{Probably omit!}
Note that there is an implicit choice made here, that propagators
already in the queue are not added.  We could choose instead to
delete them from the queue and re-add them.
}

\subsection{Basic Queuing Strategy Experiments}

\begin{table}
\caption{Queue versus stack experiments.}
\label{table:queue-stack}
\begin{center}\footnotesize
\begin{tabular}{|l||rr|rr|}
\hline
Example&\multicolumn{2}{c|}{queue}&\multicolumn{2}{c|}{stack}\\
&\multicolumn{1}{c}{time (ms)}&\multicolumn{1}{c|}{steps}&\multicolumn{1}{c}{time}&\multicolumn{1}{c|}{steps}\\\hline\hline
\texttt{all-interval-500}&$73.00$&$377\,777$&$+13626.0\%$&$+32.2\%$\\\hline
\texttt{alpha}&$97.31$&$207\,470$&$+47.5\%$&$+99.0\%$\\\hline
\texttt{bibd-7-3-60}&$2\,020.30$&$1\,020\,162$&$+7.7\%$&$+102.7\%$\\\hline
\texttt{cars}&$4.65$&$14\,860$&$+1.3\%$&$+2.6\%$\\\hline
\texttt{crowded-chess-7}&$553.25$&$660\,525$&$+4.7\%$&$+68.1\%$\\\hline
\texttt{donald-b}&$0.62$&$414$&$+36.3\%$&$+31.9\%$\\\hline
\texttt{donald-d}&$28.53$&$40$&$-5.4\%$&$+7.5\%$\\\hline
\texttt{donald-v}&$0.36$&$366$&$+7.6\%$&$+23.2\%$\\\hline
\texttt{golomb-10-b}&$1\,342.48$&$2\,095\,161$&$+870.6\%$&$+171.5\%$\\\hline
\texttt{golomb-10-d}&$3\,201.84$&$2\,091\,691$&$-11.9\%$&$+98.9\%$\\\hline
\texttt{graph-color}&$33.54$&$4\,422$&$+299.7\%$&$+118.4\%$\\\hline
\texttt{grocery}&$56.87$&$2\,211$&$+44.2\%$&$-6.3\%$\\\hline
\texttt{knights-10}&$6.66$&$25\,225$&$+7.7\%$&$+53.3\%$\\\hline
\texttt{minsort-200}&$141.84$&$113\,437$&$+1580.1\%$&$+1123.3\%$\\\hline
\texttt{o-latin-7-d}&$532.48$&$311\,135$&$+4.9\%$&$+24.9\%$\\\hline
\texttt{partition-32}&$9\,001.24$&$13\,939\,101$&$-18.5\%$&$+34.8\%$\\\hline
\texttt{photo}&$90.37$&$296\,661$&$+11.1\%$&$+8.7\%$\\\hline
\texttt{picture}&$1\,583.12$&$119\,406$&$+27.1\%$&$+79.9\%$\\\hline
\texttt{queens-400}&$549.36$&$268\,771$&$\pm 0.0\%$&$-0.1\%$\\\hline
\texttt{queens-400-a}&$14.40$&$1\,265$&$-0.9\%$&$-0.3\%$\\\hline
\texttt{sequence-500}&$98.12$&$56\,048$&$+53.8\%$&$+240.9\%$\\\hline
\texttt{square-5-d}&$32\,263.12$&$1\,478\,403$&$+71.5\%$&$+57.7\%$\\\hline
\texttt{square-7-b}&$9\,491.84$&$6\,054\,656$&$+75.3\%$&$+44.9\%$\\\hline
\texttt{square-7-v}&$5\,345.00$&$9\,636\,675$&$+8.3\%$&$+35.2\%$\\\hline
\texttt{warehouse}&$0.70$&$2\,075$&$+11.7\%$&$+29.3\%$\\\hline
\hline
average (all)& --- & --- &$+78.4\%$&$+61.0\%$\\\hline
\end{tabular}
\end{center}

\end{table}

Table~\ref{table:queue-stack} compares using a FIFO queue versus
a LIFO stack.  The result, according with folklore knowledge,
clearly illustrates that a queue must be used.  The few cases
where a stack is better are comprehensively outweighed by the
worst cases for a stack.

Later experiments in Section~\ref{sec:prioexp} using priorities
with combinations of FIFO queues and LIFO stacks reveal
that the pathological behavior of \texttt{all-interval-500} is
due to priority inversion. However, the pathological behavior of
\texttt{minsort-200} is due to the use of a LIFO stack.

\subsection{Static Priorities}

A statically prioritized queue associates with each propagator a
fixed priority, we will assume an integer in the range
$\range{0}{k-1}$.  In effect, the queue $Q$ is split into $k$
queues, $Q[0]$, \ldots $Q[k-1]$ where each $Q[i]$ is a FIFO queue
for the propagators with priority $i$.  Selection always chooses
the oldest propagator in the lowest numbered queue $Q[i]$ that
is non-empty.
Static prioritization allows one to ensure that quick propagators
are executed before slow propagators.

We give an example of seven static priorities, with names of the
integer priorities as follows:
\PPN{UNARY}=0,
\PPN{BINARY}=1,
\PPN{TERNARY}=2,
\PPN{LINEAR}=3,
\PPN{QUADRATIC}=4,
\PPN{CUBIC}=5, and
\PPN{VERYSLOW}=6.
The names are meant to represent the arity of the constraint,
and then the asymptotic runtime of the propagator, once the constraint can handle
$n$ variables. So \PPN{BINARY} is for binary constraints, \PPN{QUADRATIC} is for
constraints that are approximately $O(n^2)$ for instances with
$n$ variables.

\begin{example}[\textbf{Propagator priorities}]
For example, the propagator $f_{\cI}$ for $even(x_1)$ defined
by 
$$
\begin{array}{rcl@{\quad}l}
f_{\cI}(D)(x_1) &=& D(x_1) \cap \range{2\lceil \frac{1}{2} \inf_D x_1
\rceil}{2 \lfloor \frac{1}{2} \sup_D(x_1) \rfloor}\\ 
f_{\cI}(D)(x) &=& D(x) & x \neq x_1
\end{array}
$$
might be given priority \PPN{UNARY}, 
while
$f_{\cC}$ and $f_{\cG}$ might be given priority
\PPN{BINARY}. 

The  domain propagator defined in~\cite{Regin94}
for the \calldiff{} constraint 
$\wedge_{i=1}^{n} \wedge_{j=i+1}^n x_i \neq x_j$
(with complexity $O(n^{2.5})$) might be given priority \PPN{QUADRATIC}.

The \calldiff{} \ibounds{} propagator
defined in~\cite{puget-alldifferent} (with complexity $O(n \log n)$)
might be given priority \PPN{LINEAR}.
\end{example}

Priorities in effect force many more fixpoints to be calculated.
A fixpoint of all propagators at priority level $i$ and lower 
must be
reached before a propagator at priority level $i+1$ is run.
This means we will often cause more propagators to
run when using priorities, but more cheap propagators!

\begin{example}[\textbf{Repeated fixpoints}]\label{ex:ifix}
Consider the execution of a system of propagators
for constraints 
$c_\cR \equiv x_1 = 2 x_2$,
$c_\cS \equiv x_1 = 3 x_2$,
$c_\cT \equiv x_2 \leq 6 \rightarrow x_1 \leq x_3 + 7$
and 
$c_\cU \equiv \calldiff{[x_1,x_2,x_3,x_4,x_5]}$.
We will use the \rbounds{} propagators $f_\cR$, $f_\cS$,
$f_\cT$ for the first three constraints, and
the domain propagator, $f_\cU$, for \calldiff{} from~\cite{Regin94},
for the last constraint.
Let the  initial domain be
$D(x_1) = \range{0}{18}$,
$D(x_2) = \range{0}{9}$,
$D(x_3) = \range{0}{6}$, and
$D(x_4) = D(x_5) = \range{0}{3}$.
The priorities are \PPN{BINARY}, \PPN{BINARY}, \PPN{TERNARY}, and \PPN{QUADRATIC} respectively.
All propagators are at fixpoint. 

Suppose the domain of $x_1$ changes to
$\range{0}{17}$.
All propagators are enqued. The priority level
\PPN{BINARY} propagators are run to fixpoint (as in Example~\ref{ex:fix})
giving $D(x_1) = \range{0}{12}$,
$D(x_2) = \range{0}{6}$,
$D(x_3) = \range{0}{4}$
Then $f_\cT$ is scheduled and causes 
$D(x_1) = \range{0}{11}$.
Both $f_\cR$ and $f_\cS$ are enqued, and after executing
$f_\cR$, $f_\cS$, $f_\cR$ and $f_\cS$ the next fixpoint of
the \PPN{BINARY} priority propagators is reached:
$D(x_1) = \range{0}{6}$,
$D(x_2) = \range{0}{3}$,
$D(x_3) = \range{0}{2}$.
Then $f_\cT$ is scheduled again and causes no change.
After that, $f_\cU$ is executed, 
it reduces the domains of $D(x_1) = \range{4}{6}$
since all the values $\range{0}{3}$ are required for
the variables $x_2,x_3,x_4,x_5$.
Both $f_\cR$ and $f_\cS$ are enqued, and after executing
we reach their fixpoint $D(x_1) = \{6\}$,
$D(x_2) = \{3\}$,
$D(x_3) = \{2\}$.
Once again $f_\cT$ is executed for no change.
Then $f_\cU$ is executed once more obtaining
$D(x_4) = D(x_5) = \{0,1\}$. 
This is the overall fixpoint.
\end{example}

We can adjust the granularity of the priorities: we can have a
finer version of the above priorities with~14 priorities, each
priority above with a \PPN{LOW} and \PPN{HIGH} version.
This allows us to separate, for example, a domain consistent
propagator for the binary absolute value constraint $\operatorname{abs}(x)=y$
(\PPN{BINARY-LOW}) from a bounds consistent propagator for the
same constraint (\PPN{BINARY-HIGH}). This increased granularity
will be important in Section~\ref{sec:mult}.

Conversely, we may collapse the priorities into fewer levels, for
example into three levels
\PPN{UNARY-TERNARY}, \PPN{LINEAR-QUADRATIC},
\PPN{CUBIC-VERYSLOW}.

Another model for priorities in constraint propagation based on
composition operators is~\cite{composition}. The model, however,
runs all propagators of lower priority before switching
propagation back to propagators of higher priority. 
This model does not preempt computing a fixpoint for a low
priority. The model always completes a fixpoint for a given
priority level and only then possibly continues at a higher
priority level.

Most systems have some form of static priorities, typically using
two priority levels (for example, SICStus~\cite{sicstus},
Mozart~\cite{mozart}). The two levels are often not entirely
based on cost: in SICStus all indexicals have high priority and
all other lower priority.  While
ECL$^{i}$PS$^{e}$~\cite{eclipse,warwick} supports twelve priority
levels, its finite domain solver also uses only two priority levels
where another level is used to support constraint debugging. 

A similar, but more powerful approach is used by
Choco~\cite{choco} using seven priority levels allowing both LIFO
and FIFO traversal.

\ignore{
I now understand CHOCO's model but it's too bizarre to describe!

In particular, Choco uses a mix of priority queues and
stacks where adding of
propagators is controlled 
by the event type raised by propagation. Events of type $\fix$ are
organized into a LIFO stack and other events into a FIFO
queue. After propagation for a particular propagator has
finished, all events are processed and the dependent propagators
are put into an array of queue-like datastructure with a similar
cost 
to be rerun if necessary
}

Prioritizing particular operations during constraint propagation
is important in general. 
\ignore{
For (binary) arc consistency algorithms,
ordering heuristics for the operations performed during
propagation can reduce the total number of operations
required~\cite{order-csp}.
} 
For interval narrowing, prioritizing
constraints can avoid slow convergence, see for
example~\cite{interval-narrowing}.

The prioritizing of propagators by \emph{cost} is important,
inverting the priorities can lead to significant disadvantages.

\begin{example}[\textbf{Inverted priorities}]
Consider executing Example~\ref{ex:ifix} with inverted priorities.
We first execute $f_\cU$ then $f_\cT$ for no effect.
Then executing $f_\cR$ modifies $D(x_1)$, so each of
$f_\cU$ and $f_\cT$ are enqued and re-executed for no effect.
Then executing $f_\cS$ has the same behavior.
Overall we execute the propagators
$f_\cT$ and $f_\cU$
each at least 10 times, as opposed to 3 and 2 times respectively
in Example~\ref{ex:ifix}. 
Since they are the most expensive to execute we would expect
this to be slower (this is confirmed immediately below).
\end{example}

\subsection{Static Priority Experiments}
\label{sec:prioexp}

\begin{table}
\caption{Priority experiments with varying granularities.}
\label{table:prio}
\begin{center}\footnotesize
\begin{tabular}{|l||rr|rr|rr|}
\hline
Example&\multicolumn{2}{c|}{small}&\multicolumn{2}{c|}{medium}&\multicolumn{2}{c|}{full}\\
&\multicolumn{1}{c}{time}&\multicolumn{1}{c|}{steps}&\multicolumn{1}{c}{time}&\multicolumn{1}{c|}{steps}&\multicolumn{1}{c}{time}&\multicolumn{1}{c|}{steps}\\\hline\hline
\texttt{all-interval-500}&$+1.8\%$&$+0.3\%$&$+2.3\%$&$+0.3\%$&$+2.3\%$&$+0.3\%$\\\hline
\texttt{alpha}&$+1.2\%$&$\pm 0.0\%$&$+1.2\%$&$\pm 0.0\%$&$+1.2\%$&$\pm 0.0\%$\\\hline
\texttt{bibd-7-3-60}&$+0.1\%$&$+16.9\%$&$+0.1\%$&$+16.9\%$&$+0.2\%$&$+16.9\%$\\\hline
\texttt{cars}&$-1.4\%$&$-10.4\%$&$-1.5\%$&$-10.5\%$&$-1.8\%$&$-10.5\%$\\\hline
\texttt{crowded-chess-7}&$+8.1\%$&$+91.2\%$&$+9.0\%$&$+91.2\%$&$+8.8\%$&$+91.1\%$\\\hline
\texttt{donald-b}&$+1.1\%$&$\pm 0.0\%$&$+0.4\%$&$\pm 0.0\%$&$+0.9\%$&$\pm 0.0\%$\\\hline
\texttt{donald-d}&$-0.1\%$&$\pm 0.0\%$&$-5.6\%$&$+5.0\%$&$-5.5\%$&$+5.0\%$\\\hline
\texttt{donald-v}&$+1.6\%$&$\pm 0.0\%$&$+2.6\%$&$\pm 0.0\%$&$+2.2\%$&$\pm 0.0\%$\\\hline
\texttt{golomb-10-b}&$-33.1\%$&$+33.5\%$&$-31.3\%$&$+51.0\%$&$-31.6\%$&$+51.0\%$\\\hline
\texttt{golomb-10-d}&$-49.5\%$&$+29.2\%$&$-49.0\%$&$+46.7\%$&$-48.9\%$&$+46.7\%$\\\hline
\texttt{graph-color}&$-1.3\%$&$\pm 0.0\%$&$-1.3\%$&$-0.1\%$&$-0.9\%$&$-0.1\%$\\\hline
\texttt{grocery}&$+10.3\%$&$-9.2\%$&$-5.6\%$&$+1.9\%$&$-5.6\%$&$+1.9\%$\\\hline
\texttt{knights-10}&$-3.4\%$&$-19.5\%$&$-4.7\%$&$-19.9\%$&$-3.7\%$&$-19.9\%$\\\hline
\texttt{minsort-200}&$+0.5\%$&$\pm 0.0\%$&$+0.5\%$&$\pm 0.0\%$&$-0.2\%$&$\pm 0.0\%$\\\hline
\texttt{o-latin-7-d}&$-10.6\%$&$+4.6\%$&$-11.0\%$&$+4.6\%$&$-10.6\%$&$+4.6\%$\\\hline
\texttt{partition-32}&$-38.6\%$&$-18.2\%$&$-38.7\%$&$-17.8\%$&$-38.4\%$&$-17.8\%$\\\hline
\texttt{photo}&$+0.1\%$&$+3.4\%$&$+0.2\%$&$+3.4\%$&$-7.3\%$&$-1.8\%$\\\hline
\texttt{picture}&$-1.0\%$&$\pm 0.0\%$&$-1.5\%$&$\pm 0.0\%$&$+0.2\%$&$\pm 0.0\%$\\\hline
\texttt{queens-400}&$+0.7\%$&$\pm 0.0\%$&$+0.5\%$&$\pm 0.0\%$&$+0.5\%$&$\pm 0.0\%$\\\hline
\texttt{queens-400-a}&$+0.2\%$&$\pm 0.0\%$&$+0.2\%$&$\pm 0.0\%$&$+0.3\%$&$\pm 0.0\%$\\\hline
\texttt{sequence-500}&$+0.5\%$&$\pm 0.0\%$&$+0.3\%$&$\pm 0.0\%$&$+0.6\%$&$\pm 0.0\%$\\\hline
\texttt{square-5-d}&$-0.5\%$&$\pm 0.0\%$&$+0.9\%$&$+23.8\%$&$+1.1\%$&$+23.8\%$\\\hline
\texttt{square-7-b}&$+1.3\%$&$\pm 0.0\%$&$+0.4\%$&$\pm 0.0\%$&$-23.4\%$&$+22.0\%$\\\hline
\texttt{square-7-v}&$+0.9\%$&$\pm 0.0\%$&$+0.7\%$&$\pm 0.0\%$&$+0.9\%$&$\pm 0.0\%$\\\hline
\texttt{warehouse}&$+5.3\%$&$+13.0\%$&$+5.7\%$&$+12.0\%$&$+5.3\%$&$+12.0\%$\\\hline
\hline
average (all)&$-5.6\%$&$+3.8\%$&$-6.3\%$&$+6.4\%$&$-7.4\%$&$+7.0\%$\\\hline
\end{tabular}
\end{center}

\end{table}

\paragraph*{Priority granularity}

Table~\ref{table:prio} gives runtime and propagation steps of
various priority granularities compared to a propagation engine
using a FIFO queue (and using dynamic fixpoint reasoning, events
of types $\{\fix,\bc,\dmc\}$, and fully dynamic event sets). The
three different experiment capture different priority
granularities: ``small'' uses three priorities
(\PPN{UNARY-TERNARY}, \PPN{LINEAR-QUADRATIC},
\PPN{CUBIC-VERYSLOW}); ``medium'' uses seven priority levels
(from \PPN{UNARY} to \PPN{VERYSLOW}); ``full'' uses~14 priority
levels (from \PPN{UNARY-HIGH} to \PPN{VERYSLOW-LOW}).

\ignore{
Adding priorities has the potential to significantly increase the
number of propagations, since the engine needs to reach a
fixpoint at each priority level, but the idea is that the extra
propagations at high priority are cheap and pay for themselves by
reducing propagations at low priority.
}

The results illustrate that even when there are substantially
more propagations (for example, \texttt{golomb-10-\{b,d\}}) there
can be significant savings.  Priorities can have a very
substantial saving (almost $50\%$ for \texttt{golomb-10-d}) and
the worst case cost on the benchmarks is only 10.3\%.  Overall
while the ``medium'' range of priorities is preferable on many
benchmarks, the ``full'' range of priorities gives real speedups
on two more benchmarks (\texttt{photo} and \texttt{square-7-b})
and tends to reduce the worst case behavior of ``medium''.

Examples such as \texttt{queens-400}, \texttt{queens-400-a}, and
\texttt{sequence-500} for ``small'' only feature propagators with
the same priority (the number of propagation steps remains the
same). Hence the increase in runtime by less than $1\%$ describes
the overhead of using priorities at all.

Depending on the underlying system, increasing the granularity
also requires more memory. Gecode, as a system based on
recomputation and copying, constitutes the worst case in that each
copied node in the search tree maintains queues for all priority levels.
However, the average increase in used (not allocated as before)
memory is $+0.2\%$ for ``small'', $+0.7\%$ for ``medium'', and
$+1.5\%$ for ``full'' and hence can be neglected.

A broad spectrum of priorities will be useful for the
optimizations presented in Section~\ref{sec:mult}. Therefore it
is important that while ``full'' does not offer huge advantages
over ``medium'', it neither degrades overall performance nor requires
much memory.

\paragraph*{Priorities and stacks}

\begin{table}
\caption{Priority experiments with stacks and priorities.}
\label{table:prio-stack}
\begin{center}\footnotesize
\begin{tabular}{|l||rr|rr|rr|}
\hline
Example&\multicolumn{2}{c|}{for all}&\multicolumn{2}{c|}{for $1,2,3$-ary}&\multicolumn{2}{c|}{for $1,2$-ary}\\
&\multicolumn{1}{c}{time}&\multicolumn{1}{c|}{steps}&\multicolumn{1}{c}{time}&\multicolumn{1}{c|}{steps}&\multicolumn{1}{c}{time}&\multicolumn{1}{c|}{steps}\\\hline\hline
\texttt{all-interval-500}&$-0.6\%$&$\pm 0.0\%$&$+1.3\%$&$\pm 0.0\%$&$+0.2\%$&$\pm 0.0\%$\\\hline
\texttt{alpha}&$+47.2\%$&$+99.0\%$&$+2.3\%$&$\pm 0.0\%$&$+1.0\%$&$\pm 0.0\%$\\\hline
\texttt{bibd-7-3-60}&$+5.5\%$&$+49.7\%$&$-4.0\%$&$+1.4\%$&$-5.9\%$&$+1.4\%$\\\hline
\texttt{cars}&$-4.2\%$&$-7.3\%$&$+1.2\%$&$-2.0\%$&$+1.8\%$&$-2.0\%$\\\hline
\texttt{crowded-chess-7}&$+0.6\%$&$+10.4\%$&$+2.0\%$&$\pm 0.0\%$&$+1.7\%$&$\pm 0.0\%$\\\hline
\texttt{donald-b}&$+0.3\%$&$\pm 0.0\%$&$+1.6\%$&$\pm 0.0\%$&$+0.9\%$&$\pm 0.0\%$\\\hline
\texttt{donald-d}&$-0.6\%$&$+2.4\%$&$+1.0\%$&$\pm 0.0\%$&$+0.3\%$&$\pm 0.0\%$\\\hline
\texttt{donald-v}&$+7.7\%$&$+23.2\%$&$+2.5\%$&$\pm 0.0\%$&$+1.7\%$&$\pm 0.0\%$\\\hline
\texttt{golomb-10-b}&$+7.1\%$&$+27.0\%$&$+9.7\%$&$+27.0\%$&$+1.2\%$&$\pm 0.0\%$\\\hline
\texttt{golomb-10-d}&$+5.7\%$&$+39.2\%$&$+7.7\%$&$+39.2\%$&$+1.0\%$&$\pm 0.0\%$\\\hline
\texttt{graph-color}&$+62.7\%$&$+77.1\%$&$+2.3\%$&$\pm 0.0\%$&$+1.5\%$&$\pm 0.0\%$\\\hline
\texttt{grocery}&$+42.8\%$&$+7.1\%$&$+45.8\%$&$+7.1\%$&$+1.6\%$&$-0.3\%$\\\hline
\texttt{knights-10}&$+0.1\%$&$+5.0\%$&$+2.1\%$&$+5.0\%$&$+1.4\%$&$+5.0\%$\\\hline
\texttt{minsort-200}&$+1545.4\%$&$+1106.4\%$&$+2.2\%$&$\pm 0.0\%$&$+2.4\%$&$\pm 0.0\%$\\\hline
\texttt{o-latin-7-d}&$-3.3\%$&$+5.2\%$&$+1.0\%$&$+0.3\%$&$-0.4\%$&$+0.3\%$\\\hline
\texttt{partition-32}&$-2.2\%$&$-2.9\%$&$+2.2\%$&$+2.3\%$&$+2.2\%$&$+0.4\%$\\\hline
\texttt{photo}&$-1.3\%$&$-0.8\%$&$-1.0\%$&$-0.9\%$&$+0.1\%$&$-0.9\%$\\\hline
\texttt{picture}&$+21.1\%$&$+79.9\%$&$-1.3\%$&$\pm 0.0\%$&$-1.2\%$&$\pm 0.0\%$\\\hline
\texttt{queens-400}&$-0.1\%$&$-0.1\%$&$+1.0\%$&$-0.1\%$&$+0.8\%$&$-0.1\%$\\\hline
\texttt{queens-400-a}&$-1.0\%$&$-0.3\%$&$+1.2\%$&$\pm 0.0\%$&$+1.2\%$&$\pm 0.0\%$\\\hline
\texttt{sequence-500}&$+54.1\%$&$+240.9\%$&$+1.4\%$&$\pm 0.0\%$&$+0.8\%$&$\pm 0.0\%$\\\hline
\texttt{square-5-d}&$+65.2\%$&$+31.6\%$&$+1.2\%$&$\pm 0.0\%$&$+0.8\%$&$\pm 0.0\%$\\\hline
\texttt{square-7-b}&$+1.7\%$&$+18.6\%$&$+1.4\%$&$\pm 0.0\%$&$+0.8\%$&$\pm 0.0\%$\\\hline
\texttt{square-7-v}&$+8.9\%$&$+35.2\%$&$+2.9\%$&$\pm 0.0\%$&$+2.6\%$&$\pm 0.0\%$\\\hline
\texttt{warehouse}&$-2.5\%$&$+1.7\%$&$+3.0\%$&$+0.2\%$&$+3.0\%$&$+0.2\%$\\\hline
\hline
average (all)&$+23.9\%$&$+36.0\%$&$+3.3\%$&$+2.8\%$&$+0.8\%$&$+0.2\%$\\\hline
\end{tabular}
\end{center}

\end{table}

Table~\ref{table:prio-stack} gives the runtime and propagation
steps of using priorities together with stacks or combinations of
stacks and queues for the different priority levels. All numbers
are given relative to a propagation engine using the ``full''
priority spectrum with only queues for each priority level. All
propagation engines considered also use the full priority
spectrum. The propagation engine for ``for all'' uses only stacks
for all priority levels, whereas ``for $1,2,3$-ary'' (``for
$1,2$-ary'') uses stacks for priority levels \PPN{UNARY-HIGH}
to \PPN{TERNARY-LOW} (\PPN{UNARY-HIGH} to \PPN{BINARY-LOW})
and queues for the other levels.

The numbers for ``for all'' clarify that the misbehavior of LIFO
stacks is not due to priority inversion. The folklore belief
that LIFO stacks are good for small propagators is refuted by the
numbers for ``for $1,2,3$-ary'' and ``for $1,2$-ary''. Only three
examples show improvement in both cases while \texttt{grocery}
``for $1,2,3$-ary'' already exhibits pathological behavior. The
measurements show that queue versus stack does not matter for
unary or binary constraints whereas stacks are wrong for anything
else.

\paragraph*{Issues to avoid}

\begin{table}
\caption{Priority experiments: issues to avoid.}
\label{table:prio-wrong}
\begin{center}\footnotesize
\begin{tabular}{|l||rr|rr|}
\hline
Example&\multicolumn{2}{c|}{complete fixpoints}&\multicolumn{2}{c|}{inverse priorities}\\
&\multicolumn{1}{c}{time}&\multicolumn{1}{c|}{steps}&\multicolumn{1}{c}{time}&\multicolumn{1}{c|}{steps}\\\hline\hline
\texttt{all-interval-500}&$-0.8\%$&$\pm 0.0\%$&$+13296.5\%$&$+32.1\%$\\\hline
\texttt{alpha}&$+0.5\%$&$\pm 0.0\%$&$+0.3\%$&$\pm 0.0\%$\\\hline
\texttt{bibd-7-3-60}&$-11.7\%$&$-45.8\%$&$-7.7\%$&$-42.8\%$\\\hline
\texttt{cars}&$-1.2\%$&$-0.9\%$&$+14.4\%$&$+40.1\%$\\\hline
\texttt{crowded-chess-7}&$-4.2\%$&$-41.1\%$&$-7.4\%$&$-52.0\%$\\\hline
\texttt{donald-b}&$+0.9\%$&$\pm 0.0\%$&$+36.8\%$&$+31.9\%$\\\hline
\texttt{donald-d}&$+0.9\%$&$\pm 0.0\%$&$+6.2\%$&$-7.1\%$\\\hline
\texttt{donald-v}&$+1.6\%$&$\pm 0.0\%$&$+0.2\%$&$\pm 0.0\%$\\\hline
\texttt{golomb-10-b}&$-2.5\%$&$-14.5\%$&$+892.0\%$&$-5.7\%$\\\hline
\texttt{golomb-10-d}&$+14.6\%$&$-14.2\%$&$+945.4\%$&$-2.2\%$\\\hline
\texttt{graph-color}&$+1.9\%$&$+0.1\%$&$+336.8\%$&$+51.1\%$\\\hline
\texttt{grocery}&$+29.3\%$&$-23.4\%$&$+35.3\%$&$+9.1\%$\\\hline
\texttt{knights-10}&$-0.9\%$&$\pm 0.0\%$&$+8.7\%$&$+43.5\%$\\\hline
\texttt{minsort-200}&$+5.5\%$&$+8.6\%$&$+2732.2\%$&$+2246.7\%$\\\hline
\texttt{o-latin-7-d}&$+7.3\%$&$-3.8\%$&$+105.1\%$&$+18.7\%$\\\hline
\texttt{partition-32}&$+0.4\%$&$-1.2\%$&$+108.3\%$&$+66.8\%$\\\hline
\texttt{photo}&$+1.2\%$&$\pm 0.0\%$&$+18.6\%$&$-1.9\%$\\\hline
\texttt{picture}&$-3.6\%$&$\pm 0.0\%$&$-3.7\%$&$\pm 0.0\%$\\\hline
\texttt{queens-400}&$-0.7\%$&$\pm 0.0\%$&$+0.2\%$&$\pm 0.0\%$\\\hline
\texttt{queens-400-a}&$+0.3\%$&$\pm 0.0\%$&$-0.3\%$&$\pm 0.0\%$\\\hline
\texttt{sequence-500}&$+0.2\%$&$\pm 0.0\%$&$-0.2\%$&$\pm 0.0\%$\\\hline
\texttt{square-5-d}&$+32.5\%$&$+7.9\%$&$+50.1\%$&$+7.5\%$\\\hline
\texttt{square-7-b}&$+0.5\%$&$\pm 0.0\%$&$+177.4\%$&$-7.6\%$\\\hline
\texttt{square-7-v}&$+2.6\%$&$\pm 0.0\%$&$\pm 0.0\%$&$\pm 0.0\%$\\\hline
\texttt{warehouse}&$-3.7\%$&$-9.0\%$&$+14.7\%$&$+11.3\%$\\\hline
\hline
average (all)&$+2.5\%$&$-6.6\%$&$+107.5\%$&$+18.4\%$\\\hline
\end{tabular}
\end{center}

\end{table}

Table~\ref{table:prio-wrong} shows runtime and propagation steps
of using priorities in flawed ways. The experiment ``complete
fixpoints'' refers to the model proposed in~\cite{composition}
where fixpoints are always completed before possibly switching to
a higher priority level. As to be expected, the number of
propagation steps is reduced, however at the expense of increased
runtime. More importantly, two examples exhibiting substantial
slowdown (\texttt{golomb-10-d} and \texttt{square-5-d}) are
particularly relevant as they feature propagators of vastly
different priority levels.

The surprising behavior of \texttt{bibd-7-3-60} appears to be
due to a problem with how the priorities for the two different
kinds of propagators used in this example (\clex{} and
Boolean-sum) are classified. This clarifies that even
with a rich spectrum of priority levels at disposal it remains
difficult to assign priority levels to propagators.

The experiment ``inverse priorities'' shows numbers for a
propagation engine where a propagator with lowest priority is
executed first.  The experiment shows that there is a point to
the priority levels, high priority = fast propagator. While the
number of propagations is often reduced the approach is rarely
better than no priorities and sometimes catastrophically worse.

The experiment ``inverse priorities'' clarifies a very important
aspect of priorities: they not only serve as a means to improve
performance, they also serve as a \emph{safeguard} against
pathological propagation order.

\subsection{Dynamic Priorities}

As evaluation proceeds, variables become fixed and propagators
can be replaced by more specialized versions.  If a propagator is
replaced by a more specialized version, also its priority should
change.

\begin{example}[\textbf{Updating a propagator}]
Consider the propagator $f_{\cJ}$ for updating
$x_1$ in the constraint $x_1 = x_2 + x_3$
defined by 
$$
\begin{array}{lcl@{\quad}l}
f_{\cJ}(D)(x_1) & = & D(x_1) \cap \range{\inf_D(x_2) + \inf_D(x_3)}{
                               \sup_D(x_2) + \sup_D(x_3)} \\
f_{\cJ}(D)(x)   & = & D(x) & x \neq x_1
\end{array}
$$
might have initial priority \PPN{TERNARY}.
When the variable $x_2$ becomes fixed to $d_2$ say, then the implementation
for $x_1$ can change to 
$$\textstyle
f_{\cJ}(D)(x_1)  =  D(x_1) \cap \range{d_2 + \inf_D(x_3)}{
                               d_2 + \sup_D(x_3)}
$$
and the priority can change to \PPN{BINARY}.
\end{example}

Changing priorities is also relevant when a propagator with $n>3$
variables with priority \PPN{LINEAR} (or worse) reduces to a
binary or ternary propagator.

\subsection{Dynamic Priority Experiments}

\begin{table}
\caption{Dynamic priority experiments.}
\label{table:prio-dynamic}
\begin{center}\footnotesize
\begin{tabular}{|l||rr|}
\hline
Example&\multicolumn{2}{c|}{dynamic}\\
&\multicolumn{1}{c}{time}&\multicolumn{1}{c|}{steps}\\\hline\hline
\texttt{alpha}&$-25.5\%$&$-41.7\%$\\\hline
\texttt{cars}&$-5.9\%$&$-13.4\%$\\\hline
\texttt{crowded-chess-7}&$-5.0\%$&$-26.5\%$\\\hline
\texttt{o-latin-7-d}&$-7.3\%$&$-7.4\%$\\\hline
\texttt{picture}&$-1.9\%$&$\pm 0.0\%$\\\hline
\texttt{sequence-500}&$+6.0\%$&$+34.8\%$\\\hline
\hline
average (above)&$-7.1\%$&$-12.1\%$\\\hline
average (all)&$-1.7\%$&$-3.0\%$\\\hline
\end{tabular}
\end{center}

\end{table}

Table~\ref{table:prio-dynamic} shows runtime and propagation
steps for an engine using dynamic priorities compared to an
engine using all optimizations introduced earlier and the full
priority spectrum.  Dynamically changing the priority of
propagators as they become smaller due to fixed variables can
lead to significant improvements.  In effect, constraints that
become smaller (and thus run at higher priority) are run first
causing the still large constraints to be run less often.

This is in particular true for \texttt{alpha} with initially only
propagators for linear equalities with priority
\PPN{LINEAR}. When fixing variables during search many of these
propagators are then run at priority levels \PPN{BINARY} and
\PPN{TERNARY}. It is worth noting that using dynamic priorities
can disturb the FIFO queue behavior: for \texttt{sequence-500}
it appears to be more important to run all \cexactly{} propagators at the
same priority level. Running some of the \cexactly{} propagators
at priorities \PPN{BINARY} and \PPN{TERNARY} is not beneficial
and disturbs the queue behavior.

Dynamic priorities incur the overhead to compute the priority
based on the number of not yet fixed variables. However, the
overhead is still small enough to make dynamic priorities
worthwhile overall.

\section{Combining Propagation}
\label{sec:mult}

There are many ways to define a correct propagator $f$
for a single constraint $c$: the art of building propagators
is to find good tradeoffs in terms of speed of execution
versus strength of propagation. 
Typically a single constraint may have a number of different
propagator implementations: the cheapest simple propagator,
a more complex bounds propagator, and a more complex domain propagator,
for example.

\begin{example}[\textbf{\calldiff{} propagators}]\label{ex:mult}
Consider the propagator $f_{\cK}(D)$
for the \calldiff{} constraint.
\begin{quote}
\begin{tabbing}
xx \= xx \= xx \= xx \= xx \= \kill
\> $E$ := $\emptyset$ \\
\> \textbf{for}  $i \in \range{1}{n}$ \\
\> \> \textbf{if} ($\exists d. D(x_i) = \{d\}$) \\
\> \> \> \textbf{if} ($d \in E$) \textbf{return} $D_\bot$
\textbf{else} $E$ := $E \cup \{d\}$ \\
\> \textbf{for} $i \in \range{1}{n}$ \\
\> \> \textbf{if} ($|D(x_i)| > 1$) $D(x_i)$ := $D(x_i) - E$ \\
\> \textbf{return} $D$
\end{tabbing}
\end{quote}
The propagator does a linear number of set operations in each
invocation and is checking. It can be made idempotent by testing
that no variable becomes fixed.

Another propagator for the same constraint is the domain
propagator $f_{\cL}$ introduced in~\cite{Regin94} with complexity
$O(n^{2.5})$.
\end{example}

Given two propagators, say $f_1$ and $f_2$ in $\prop(c)$, where
$f_1$ is strictly stronger than $f_2$ ($f_1(D) \sqsubseteq
f_2(D)$ for all domains $D$), we could choose to implement $c$ by
just $f_1$ or just $f_2$ trading off pruning versus execution
time.  

Without priorities there is no point in implementing the
constraint $c$ using both propagators, since $f_1$ will always be
run and always compute stronger domains than $f_2$.  

We could possibly merge the implementation of the propagators to
create a new propagator $f_{12}(D) = f_1(f_2(D))$.  By running
the cheaper propagator immediately first we hope that we can (a)
quickly determine failure in some cases and (b) simplify the
domains before applying the more complicated propagator $f_1$.
While this immediate combination of two propagators in essence is
simply building a new propagator, once we have priorities in our
propagation engine we can use two or more propagators for the
same constraint in different ways.

\subsection{Multiple Propagators}
\label{subsec:multi}

Once we have priorities it makes sense to use multiple
propagators to implement the same constraint.  We can run the
weaker (and presumably faster) propagator $f_2$ with a higher
priority than $f_1$. This makes information available earlier to
other propagators.  When the stronger propagator $f_1$ is
eventually run, it is able to take advantage from propagation
provided by other cheaper propagators.

Note that this is essentially different from having a single
propagator $f_{12}$ that always first runs the algorithm of $f_2$
and then the algorithm of $f_1$.

\begin{example}[\textbf{Multiple \calldiff{}}]\label{ex:mult2}
  Consider the two propagators $f_\cK$ and $f_\cL$ defined in
  Example~\ref{ex:mult} above.  We can use both propagators:
  $f_{\cK}$ with priority \PPN{LINEAR}, and $f_{\cL}$ with
  priority \PPN{QUADRATIC}. This means that we will not invoke
  $f_{\cL}$ until we have reached a fixpoint of $f_{\cK}$ and all
  \PPN{LINEAR} and higher priority propagators.

Consider the additional propagator $f_{\cC}$ for the
constraint $3x_1 = 2x_2$, which has priority \PPN{BINARY}. 
Consider the domain $D$ where
$D(x_1) = \{4,6\}$,  $D(x_2) = \{6,9\}$, $D(x_3) = \{6,7\}$
and $D(x_4) = \cdots = D(x_n) = \range{1}{n}$, which is
a fixpoint for $f_{\cC}, f_{\cK}$ and $f_{\cL}$. 
Now assume the domain of $x_3$ is reduced to 6.
Propagator $f_{\cK}$ is placed in queue \PPN{LINEAR} and
$f_{\cL}$ is placed in queue \PPN{QUADRATIC}.
Applying $f_{\cK}$ removes 6 from the domain
of all the domains of $x_1, x_2, x_4, \ldots, x_n$, and this
causes $f_{\cC}$ to be placed in queue \PPN{BINARY}.
This is the next propagator considered and it causes failure.
Propagator $f_{\cL}$ is never executed.

If we just use $f_{\cL}$ then we need to invoke the
more expensive $f_{\cL}$ to obtain the same domain changes as
$f_{\cK}$, and then fail.
\end{example}

\subsection{Staged Propagators}
\label{subsec:staged}

Once we are willing to use multiple propagators for a single
constraint it becomes worth considering how to more efficiently
manage them.  Instead of using two (or more) distinct propagators
we can combine the several propagators into a single propagator
with more effective behavior.

We assume that a propagator has an internal state variable,
called its \emph{stage}.  When it is invoked, the stage
determines what form of propagation applies.

\begin{example}[\textbf{Staged \calldiff{}}]
Consider the \calldiff{} constraint with
implementations $f_{\cK}$ and $f_{\cL}$ discussed in
Example~\ref{ex:mult}. 
We combine them into a staged propagator as follows:
\begin{longitem}
\item On a $\fix(x)$ event, the propagator is moved to stage~A,
and placed in the queue with priority \PPN{LINEAR}.
\item On a $\dmc(x)$ event, unless the propagator is in stage~A already,
the propagator is put in stage~B, and placed in the queue with 
priority \PPN{QUADRATIC}.
\item Execution in stage~A uses $f_{\cK}$, 
the propagator is put in stage~B, and placed in the queue  with 
priority \PPN{QUADRATIC}, unless it is subsumed.
\item Execution in stage~B uses $f_{\cL}$, afterwards
  the propagator is removed from all queues (stage NONE).
\end{longitem}

The behavior of the staged propagator is identical to
the multiple propagators for the sample execution of
Example~\ref{ex:mult}.
In addition to the obvious advantage of having a single staged
propagator, another advantage comes from avoiding the execution of
$f_{\cL}$ when the constraint is subsumed.
\end{example}

In addition to giving other propagators with higher priority the
opportunity to run before the expensive part of a staged
propagator, the first stage of a propagator can already determine
that the next second does not need to be run. This is illustrated
by the following example.

\begin{example}[\textbf{Staged linear equations}]
  \label{ex:staged:fix}%
  Consider the unit coefficient linear equation $\Sigma_{i=1}^n
  a_i x_i = d$ constraint where $|a_i| = 1, 1 \leq i \leq n$.  We
  have two implementations, $f_{\cM}$, which implements
  \rbounds{} consistency (considering real solutions, with linear
  complexity) for the constraint, and $f_{\cN}$, which implements
  domain consistency (with exponential complexity).
  
  We combine them into a staged propagator as follows:
\begin{longitem}
\item On a $\bc(x)$ (or depending on the event types available:
  $\lbc(x)$ or $\ubc(x)$) event, the propagator is moved to
  stage~A, and placed in the queue with priority \PPN{LINEAR}.
\item On a $\dmc(x)$ event, unless the propagator is in stage A
  already, the propagator is put in stage B, and is placed in the
  queue with priority \PPN{VERYSLOW}.
\item Execution in stage~A uses $f_{\cM}$, afterwards the
  propagator is put in stage~B, and placed in the queue with
  priority \PPN{VERYSLOW}, unless each $x_i$ has a range domain
  in which case it is removed from all queues (stage NONE).
\item Execution in stage~B uses $f_{\cN}$, afterwards the
  propagator is removed from all queues (stage NONE).
\end{longitem}

The staged propagator is advantageous since the ``fast''
propagator $f_{\cM}$ can more often determine that its result $D'
= f_{\cM}(D)$ is also a fixpoint for $f_{\cN}$.
\end{example}

Staged propagators are widely applicable. They can be used similarly for the $\ibounds$
version of the \calldiff{} constraint. Another area where staged
propagators can be used is constraint-based scheduling, where
typically different propagation methods with different strength
and efficiency are available~\cite{scheduling}. 

Staging is not limited to expensive propagators, it is already
useful for binary (for example, combining bounds and domain
propagation for the absolute value constraint
$\operatorname{abs}(x)=y$) and ternary constraints (for example,
combining bounds and domain propagation for the multiplication
constraint $x\times y=z$).

It is important to note that staging requires a sufficiently
rich spectrum of priorities. For example, to use staging for
binary or ternary propagators as mentioned above, at least two
different priority levels must be available for staging. This
explains why the full priority spectrum is useful: here, for
binary propagators two priorities \PPN{BINARY-HIGH} and
\PPN{BINARY-LOW} are available. Likewise, \PPN{TERNARY-HIGH} and
\PPN{TERNARY-LOW} are available for ternary propagators.

\subsection{Combining Propagation Experiments}

\begin{table}
\caption{Combination experiments.}
\label{table:comb}
\begin{center}\footnotesize
\begin{tabular}{|l||rr|rr|rr|}
\hline
Example&\multicolumn{2}{c|}{immediate}&\multicolumn{2}{c|}{multiple}&\multicolumn{2}{c|}{staged}\\
&\multicolumn{1}{c}{time}&\multicolumn{1}{c|}{steps}&\multicolumn{1}{c}{time}&\multicolumn{1}{c|}{steps}&\multicolumn{1}{c}{time}&\multicolumn{1}{c|}{steps}\\\hline\hline
\texttt{donald-b}&$+0.7\%$&$\pm 0.0\%$&$-15.4\%$&$+10.4\%$&$-16.9\%$&$+10.1\%$\\\hline
\texttt{donald-d}&$-1.6\%$&$-2.4\%$&$\pm 0.0\%$&$+111.9\%$&$-1.6\%$&$+88.1\%$\\\hline
\texttt{golomb-10-b}&$+0.3\%$&$\pm 0.0\%$&$-10.1\%$&$+9.2\%$&$-9.8\%$&$+9.2\%$\\\hline
\texttt{golomb-10-d}&$-2.7\%$&$+0.2\%$&$-14.5\%$&$+8.6\%$&$-16.1\%$&$+8.6\%$\\\hline
\texttt{graph-color}&$-0.6\%$&$-0.6\%$&$+2.0\%$&$+23.0\%$&$-14.4\%$&$+12.3\%$\\\hline
\texttt{o-latin-7-d}&$-1.2\%$&$+1.0\%$&$-9.6\%$&$+14.3\%$&$-14.0\%$&$+10.9\%$\\\hline
\texttt{partition-32}&$-0.6\%$&$\pm 0.0\%$&$-4.3\%$&$+7.4\%$&$-6.2\%$&$+4.2\%$\\\hline
\texttt{photo}&$\pm 0.0\%$&$\pm 0.0\%$&$-0.9\%$&$+7.8\%$&$-1.5\%$&$+7.8\%$\\\hline
\texttt{picture}&$+0.3\%$&$\pm 0.0\%$&$-0.1\%$&$\pm 0.0\%$&$-3.4\%$&$\pm 0.0\%$\\\hline
\texttt{square-5-d}&$-29.6\%$&$-7.4\%$&$-38.3\%$&$+28.8\%$&$-42.8\%$&$+56.0\%$\\\hline
\texttt{square-7-b}&$-0.1\%$&$\pm 0.0\%$&$-17.5\%$&$-11.0\%$&$-18.6\%$&$-10.9\%$\\\hline
\hline
average (above)&$-3.6\%$&$-0.9\%$&$-10.7\%$&$+16.1\%$&$-14.0\%$&$+15.3\%$\\\hline
average (all)&$-1.6\%$&$-0.4\%$&$-4.8\%$&$+6.8\%$&$-6.5\%$&$+6.5\%$\\\hline
\end{tabular}
\end{center}

\end{table}

Table~\ref{table:comb} presents runtime and propagation steps of
different propagator combination schemes compared to a
propagation engine using the full priority spectrum and all
optimizations presented so far. The experiment ``immediate'' uses
a single propagator that always runs the first stage immediately
followed by the second stage. For experiment ``multiple'',
multiple propagators for different stages (as discussed in
Section~\ref{subsec:multi}) are used, whereas for experiment
``staged'' full staging is used (as described in
Section~\ref{subsec:staged}).

A quite surprising result is that ``immediate'' offers only
modest or even no speedup. The only exception is
\texttt{square-5-d} using \rbounds{} propagation immediately
before domain propagation for several linear equation
propagators.

Using multiple propagators leads to an average reduction in
runtime by $10\%$ with a slowdown for just a single example
(\texttt{graph-color}). As to be expected, the number of
propagation steps rises sharply. This is due to the fact that
more propagators need to be run and that, similar to the
introduction of priorities, propagators with high priority are
run more often. The exceptional case of \texttt{square-7-b} where
the number of propagation steps decreases appears to be a fortunate
change in propagation order.

Staged propagation offers another level of improvement over using
multiple propagators: all examples now achieve speedup. As fewer
propagators must be executed compared to ``multiple'', also the
number of propagation steps decreases (apart from
\texttt{square-5-d} being another case of changing
propagation order). Due to the reduced overhead compared to
``multiple'', even small examples such as \texttt{graph-color} are
able to benefit from staged execution.

The exact improvement in runtime of \texttt{donald-d} for
``immediate'' and ``staged'' is due to early fixpoint detection
for a big linear equation propagator as discussed in
Example~\ref{ex:staged:fix}.

The memory requirements for ``immediate'' and ``staged'' are
unchanged. The use of multiple propagators for ``multiple'' leads
to an average increase of $6.4\%$ in allocated memory for the
examples shown in Table~\ref{table:comb}.

In summary, staged propagation is very effective for all
examples, so it should clearly be used.

\section{Experiment Summary}
\label{sec:summary}

\begin{table}
\caption{Experiment summary.}
\label{table:monty}
\begin{center}\footnotesize
\begin{tabular}{|l||rr|rr|}
\hline
Example&\multicolumn{2}{c|}{no optimizations}&\multicolumn{2}{c|}{all optimizations}\\
&\multicolumn{1}{c}{time (ms)}&\multicolumn{1}{c|}{mem (KB)}&\multicolumn{1}{c}{time}&\multicolumn{1}{c|}{memory}\\\hline\hline
\texttt{all-interval-500}&$118.31$&$385.4$&$-36.1\%$&$+25.0\%$\\\hline
\texttt{alpha}&$106.56$&$22.2$&$-31.2\%$&$+9.8\%$\\\hline
\texttt{bibd-7-3-60}&$2\,279.36$&$6\,688.6$&$-11.0\%$&$+5.8\%$\\\hline
\texttt{cars}&$4.64$&$41.7$&$-7.7\%$&$+1.2\%$\\\hline
\texttt{crowded-chess-7}&$624.25$&$196.9$&$-8.1\%$&$-6.4\%$\\\hline
\texttt{donald-b}&$0.70$&$7.4$&$-25.7\%$&$+19.0\%$\\\hline
\texttt{donald-d}&$30.37$&$3.2$&$-13.0\%$&$+68.2\%$\\\hline
\texttt{donald-v}&$0.38$&$5.4$&$-3.3\%$&$+44.4\%$\\\hline
\texttt{golomb-10-b}&$1\,347.48$&$40.0$&$-38.5\%$&$+9.3\%$\\\hline
\texttt{golomb-10-d}&$2\,430.00$&$37.0$&$-43.7\%$&$+4.8\%$\\\hline
\texttt{graph-color}&$35.87$&$832.4$&$-19.9\%$&$+9.0\%$\\\hline
\texttt{grocery}&$55.41$&$7.7$&$-5.9\%$&$+7.9\%$\\\hline
\texttt{knights-10}&$7.46$&$770.3$&$-14.7\%$&$+4.2\%$\\\hline
\texttt{minsort-200}&$342.48$&$32\,454.5$&$-58.8\%$&$-5.1\%$\\\hline
\texttt{o-latin-7-d}&$574.36$&$242.9$&$-32.5\%$&$+5.5\%$\\\hline
\texttt{partition-32}&$8\,571.24$&$160.4$&$-40.4\%$&$+9.2\%$\\\hline
\texttt{photo}&$108.87$&$37.0$&$-23.8\%$&$+10.1\%$\\\hline
\texttt{picture}&$1\,553.42$&$450.5$&$-3.0\%$&$+18.0\%$\\\hline
\texttt{queens-400}&$4\,433.12$&$30\,286.1$&$-87.6\%$&$+0.2\%$\\\hline
\texttt{queens-400-a}&$16.15$&$566.3$&$-10.3\%$&$+4.4\%$\\\hline
\texttt{sequence-500}&$517.96$&$6\,081.5$&$-79.8\%$&$-49.4\%$\\\hline
\texttt{square-5-d}&$33\,391.24$&$43.5$&$-44.2\%$&$+11.9\%$\\\hline
\texttt{square-7-b}&$10\,166.24$&$160.3$&$-41.2\%$&$+8.7\%$\\\hline
\texttt{square-7-v}&$5\,690.00$&$144.2$&$-3.9\%$&$+11.7\%$\\\hline
\texttt{warehouse}&$0.74$&$29.4$&$+1.2\%$&$+1.0\%$\\\hline
\hline
average (all)& --- & --- &$-33.3\%$&$+7.2\%$\\\hline
\end{tabular}
\end{center}

\end{table}

In Table~\ref{table:monty} we summarize the effect of all
improvements suggested in this paper. The naive propagation
engine is compared to an engine featuring all techniques
introduced in this paper: dynamic fixpoint reasoning, $\{ \dmc, \fix,
\bc \}$ fully dynamic events, dynamic priority based LIFO
queuing with the full priority spectrum, and staged propagators.

It is interesting to note that all examples but
\texttt{warehouse} show an improvement in runtime and that almost
$75\%$ of the examples show an improvement of at least
$10\%$. The improvement in runtime does not incur a large
increase in memory: the largest increases are for the
three \texttt{donald-*} problems, where the increase is actually
negligible in absolute terms and due to the underlying memory
allocation strategy (as discussed in Section~\ref{sec:event-exp}).

The effects of the individual optimizations discussed in this
paper could be summarized as follows.  Dynamic fixpoint reasoning
subsumes static reasoning and is easy to implement, it provides
a modest improvement in execution times.  Events, while used in
all finite domain propagation engines, have less benefit than
perhaps was assumed by developers.  Using dynamic events again
leads to a modest improvement in execution times.  The fairness
of a FIFO queue strategy is essential for scheduling propagators.
While priorities by themselves are not that important they
provide a protection against worst case behavior and enable the
use of multiple propagators.  Staging is an important
optimization that can significantly improve performance.

\section{Conclusion and Future Work}
\label{sec:conclusion}

We have given a formal definition of propagation systems
including idempotence, events, and priorities used in current
propagation systems and have evaluated their impact.  We have
introduced dynamically changing event sets which are shown to
improve efficiency considerably.  The paper has introduced
multiple and staged propagators which are shown to be an
important optimization in particular for improving the efficiency of
costly global constraints.

While the improvements to an engine of a propagation based
constraint solver have been discussed for integer constraints,
the techniques readily carry over to arbitrary constraint domains
such as finite sets and multisets. 

A rather obvious way to further speed up constraint propagation
is to consider not only cost but also estimated impact for a
propagator. However, while computing cost is straightforward
it is currently not clear to us how to accurately predict
propagation impact.

\appendix

\section{Examples Used in Experiments}
\label{sec:examples}

All variants of constraint propagation discussed in the paper are
experimentally evaluated. The characteristics of the examples
used in evaluation are summarized in Table~\ref{table:examples}.
The column ``variables'' gives the number of variables in the
example, whereas the column ``propagators'' shows the number of
propagators as implementations of constraints in the example. The
column ``search'' shows which search strategy is used to search
for a solution (``first'' is simple backtracking search for the
first solution, ``all'' is search for all solutions, ``best'' is
branch-and-bound search for a best solution). The two last
columns describe how many failed nodes are explored during search
(column ``failures'') and how many solutions are found (column
``solutions'').

\begin{table}
\caption{Example characteristics.}
\label{table:examples}
\begin{center}\footnotesize
\begin{tabular}{|l||r|r||c|r|r|}
\hline
Example&\multicolumn{1}{c|}{variables}&\multicolumn{1}{c|}{propagators}&\multicolumn{1}{c|}{search}&\multicolumn{1}{c|}{failures}&\multicolumn{1}{c|}{solutions}\\
\hline
\hline
\texttt{all-interval-500}&$1\,498$&$1\,002$&first&$0$&$1$\\\hline
\texttt{alpha}&$26$&$21$&all&$7\,435$&$1$\\\hline
\texttt{bibd-7-3-60}&$11\,760$&$9\,693$&first&$1\,306$&$1$\\\hline
\texttt{cars}&$60$&$93$&all&$107$&$6$\\\hline
\texttt{crowded-chess-7}&$163$&$275$&first&$30\,396$&$1$\\\hline
\texttt{donald-b}&$10$&$2$&first&$79$&$1$\\\hline
\texttt{donald-d}&$10$&$2$&first&$5$&$1$\\\hline
\texttt{donald-v}&$10$&$2$&first&$79$&$1$\\\hline
\texttt{golomb-10-b}&$46$&$46$&best&$19\,929$&$10$\\\hline
\texttt{golomb-10-d}&$46$&$46$&best&$19\,929$&$10$\\\hline
\texttt{graph-color}&$201$&$566$&first&$37$&$1$\\\hline
\texttt{grocery}&$7$&$7$&first&$37$&$1$\\\hline
\texttt{knights-10}&$2\,028$&$2\,981$&first&$2$&$1$\\\hline
\texttt{minsort-200}&$399$&$398$&first&$0$&$1$\\\hline
\texttt{o-latin-7-d}&$147$&$133$&first&$2\,188$&$1$\\\hline
\texttt{partition-32}&$128$&$134$&first&$160\,258$&$1$\\\hline
\texttt{photo}&$61$&$54$&best&$6\,995$&$7$\\\hline
\texttt{picture}&$625$&$50$&first&$3\,242$&$1$\\\hline
\texttt{queens-400}&$400$&$239\,400$&first&$10$&$1$\\\hline
\texttt{queens-400-a}&$400$&$3$&first&$10$&$1$\\\hline
\texttt{sequence-500}&$500$&$502$&all&$250$&$1$\\\hline
\texttt{square-5-d}&$25$&$15$&first&$41\,272$&$1$\\\hline
\texttt{square-7-b}&$49$&$19$&first&$245\,208$&$1$\\\hline
\texttt{square-7-v}&$49$&$19$&first&$481\,301$&$1$\\\hline
\texttt{warehouse}&$81$&$76$&best&$20$&$4$\\\hline
\end{tabular}
\end{center}

\end{table}

A \texttt{-d} at the end of the example name means that domain
propagation is used for all occurring \calldiff{} and linear
equation constraints. Likewise, \texttt{-b} means that \ibounds{}
propagation is used for all \calldiff{} and \rbounds{} for all
linear equation constraints. In contrast, for \texttt{-v}
\rbounds{} propagation is used for all linear constraints,
whereas naive propagation (eliminating assigned values as in
Example~\ref{ex:mult}) is used for \calldiff{}.

If not otherwise mentioned, bounds consistency is used for
arithmetic constraints (including linear constraints) and naive
propagation for \calldiff{}.

\begin{longitem}
\item \texttt{all-interval-500} computes a series of numbers
  where the distances between adjacent numbers are pairwise
  distinct \csplib{007}. The model uses a single \ibounds{}
  consistent \calldiff{} propagator and many binary absolute value
  ($\operatorname{abs}(x)=y$) and ternary minus propagators.
\item \texttt{alpha} and \texttt{donald} are crypto-arithmetic
  puzzles involving linear equation propagators and a single
  \calldiff{} propagator.
\item \texttt{bibd-7-3-60} is an instance of a balanced
  incomplete block design problem with parameters
  $(v,k,l)=(7,3,60)$~\csplib{028}. The model involves Boolean-sum
  propagators and \clex{} propagators for symmetry breaking.
\item \texttt{cars} models the well known car sequencing problem
  from~\cite{exactly} using \celement, \cexactly, and linear
  equation propagators \csplib{001}.
\item \texttt{crowded-chess-7} places several different chess
  pieces on a $7\times 7$ chessboard~\cite{dudeney}. It uses
  \cexactly, \celement, domain consistent \calldiff, and
  \rbounds{} consistent linear equation propagators.
\item \texttt{golomb-10} finds an optimal Golomb ruler of
  size~$10$~\csplib{006} with the usual model. 
\item \texttt{graph-color} performs clique-based graph coloring
  for a graph with $200$ nodes. Coloring each clique
  uses a domain consistent \calldiff{} propagator.
\item \texttt{grocery} is a small crypto-arithmetic puzzle using
  in particular \rbounds{} consistent multiplication propagators.
\item \texttt{knights-10} finds a sequence of
  knight moves on a $10\times 10$ chess board such that each
  field is visited exactly once and that the moves return the
  knight to the starting field. The model uses a naive
  \calldiff{} propagator and a large number of reified binary
  propagators. 
\item \texttt{minsort-200} sorts~$200$ variables using~$200$
  minimum propagators.
\item \texttt{o-latin-7} finds an orthogonal latin square of
  size~$7$ and mostly uses domain consistent \calldiff{}
  propagators. 
\item \texttt{partition-32} partitions two~$32$ number blocks
  such that their products match. Uses several \rbounds{}
  multiplication propagators, a single domain consistent
  \calldiff{} propagator, and few linear equation propagators.
\item \texttt{photo} places $9$ persons on a picture such that as
  many preferences as possible are satisfied. Uses a large \ibounds{}
  consistent \calldiff{} propagator, a large \rbounds{} consistent
  linear propagator, and many reified binary propagators.
\item \texttt{picture} models a $25\times 25$ picture-puzzle
  \csplib{012} using $50$ \cregular{} propagators.
\item \texttt{queens-400} and \texttt{queens-400-a} places $400$
  queens on a $400\times 400$ chess board such that the queens do
  not attack each other. \texttt{queens-400} uses quadratically
  many binary disequality propagators, while
  \texttt{queens-400-a} uses three naive \calldiff-propagators.
\item \texttt{sequence-500} computes a magic sequence with $500$
  elements using $500$ \cexactly{} propagators \csplib{019}.
\item \texttt{square-5} (\texttt{square-7}) computes a magic
  square of size $5\times 5$ ($7\times 7$) using linear equation
  propagators and a single \calldiff{} propagator \csplib{019}.
\item \texttt{warehouse} solves a warehouse location problem
  following~\cite{OPL:1999}.
\end{longitem}

\section{Evaluation Platform}
\label{sec:platform}

All experiments use Gecode, a \CPP-based constraint programming
library~\cite{gecode}. Gecode is one of the fastest constraint
programming systems currently available, benchmarks comparing
Gecode to other systems are available from Gecode's webpage. The
version used in this paper corresponds to Gecode~1.3.0 (albeit
slightly modified to ease the numerous experiments in this
paper). Gecode has been compiled with Microsoft Visual Studio
Express Edition~2005.

All examples have been run on a Laptop with a $2$~GHz Pentium~M
CPU and 1024~MB main memory running Windows~XP. Runtimes are the
average of 25~runs with a coefficient of deviation less
than~$4\%$ for all benchmarks.

\begin{acks}
Christian Schulte is partially funded by the Swedish Research
Council (VR) under grant~621-2004-4953.  We thank Mikael
Lagerkvist and Guido Tack for many helpful suggestions that
improved the paper.
\end{acks}

\bibliographystyle{plain}
\bibliography{paper}

\begin{thebibliography}{10}

\bibitem{apt}
Krzysztof Apt.
\newblock {\em Principles of Constraint Programming}.
\newblock Cambridge University Press, Cambridge, United Kingdom, 2003.

\bibitem{scheduling}
Philippe Baptiste, Claude {Le Pape}, and Wim Nuijten.
\newblock {\em Constraint-based Scheduling}.
\newblock Kluwer Academic Publishers, Dordrecht, The Netherlands, 2001.

\bibitem{TRICS:00}
Nicolas Beldiceanu, Warwick Harvey, Martin Henz, François Laburthe, Eric
  Monfroy, Tobias Müller, Laurent Perron, and Christian Schulte.
\newblock Proceedings of {TRICS}: {T}echniques fo{R} {I}mplementing
  {C}onstraint programming {S}ystems, a post-conference workshop of {CP 2000}.
\newblock Technical Report TRA9/00, School of Computing, National University of
  Singapore, September 2000.

\bibitem{BenhamouHeterogeneous}
Frederic Benhamou.
\newblock {H}eterogeneous {C}onstraint {S}olving.
\newblock In {\em Proceedings of the Fifth International Conference on
  Algebraic and Logic Programming}, volume 1139 of {\em LNCS}, pages 62--76,
  Aachen, Germany, 1996. Springer-Verlag.

\bibitem{lex}
Mats Carlsson and Nicolas Beldiceanu.
\newblock Revisiting the lexicographic ordering constraint.
\newblock Technical Report T2002-17, Swedish Institute of Computer Science,
  Stockholm, Sweden, 2002.

\bibitem{CarlssonOttossonEa:97}
Mats Carlsson, Greger Ottosson, and Björn Carlson.
\newblock An open-ended finite domain constraint solver.
\newblock In {\em Programming Languages: Implementations, Logics, and Programs,
  9th International Symposium, PLILP'97}, volume 1292 of {\em LNCS}, pages
  191--206, Southampton, United Kingdom, September 1997. Springer-Verlag.

\bibitem{element}
Andre Chamard, Annie Fischler, Dominique-Benoit Guinaudeau, and Andre Guillard.
\newblock {CHIC} lessons on {CLP} methodology.
\newblock Technical report, Dassault Aviation, 1995.

\bibitem{bounds-corr}
Chiu~Wo Choi, Warwick Harvey, Jimmy Ho-Man Lee, and Peter~J. Stuckey.
\newblock Finite domain bounds consistency revisited.
\newblock Technical report, \texttt{http://arxiv.org/abs/cs.AI/0412021}, 2004.

\bibitem{CodognetDiaz:96}
Philippe Codognet and Daniel Diaz.
\newblock Compiling constraints in {{\tt clp(FD)}}.
\newblock {\em Journal of Logic Programming}, 27(3):185--226, June 1996.

\bibitem{CSPLIB}
CSPLib.
\newblock {CSPLib}: a problem library for constraints, 2006.
\newblock Available from \texttt{http://www.csplib.org}.

\bibitem{dudeney}
Henry~E. Dudeney.
\newblock {\em Amusements in Mathematics}.
\newblock Dover, New York, NY, USA, 1958.

\bibitem{gecode}
{Gecode Team}.
\newblock Gecode: Generic constraint development environment, 2006.
\newblock Available from \texttt{http://www.gecode.org}.

\bibitem{minion:wl}
Ian~P. Gent, Chris Jefferson, and Ian Miguel.
\newblock Watched literals for constraint propagation in {Minion}.
\newblock In Frédéric Benhamou, editor, {\em Twelfth International Conference
  on Principles and Practice of Constraint Programming}, volume 4204 of {\em
  LNCS}, pages 182--197. Springer-Verlag, Nantes, France, September 2006.

\bibitem{composition}
Laurent Granvilliers and Eric Monfroy.
\newblock Implementing constraint propagation by composition of reductions.
\newblock In {\em Logic Programming: Proceedings of the $19^{\textup{th}}$
  International Conference}, volume 2916 of {\em LNCS}, pages 300--314, Mumbai,
  India, 2003. Springer-Verlag.

\bibitem{warwick}
Warwick Harvey.
\newblock Personal communication, April 2004.

\bibitem{form-cons}
Warwick Harvey and Peter~J. Stuckey.
\newblock Improving linear constraint propagation by changing constraint
  representation.
\newblock {\em Constraints}, 8(2):173--207, 2003.

\bibitem{ILOG}
{ILOG S.A.}
\newblock {\em {ILOG} {Solver} {5.0}: Reference Manual}.
\newblock Gentilly, France, August 2000.

\bibitem{sicstus}
{Intelligent Systems Laboratory}.
\newblock {SICStus} {Prolog} user's manual, 3.11.1.
\newblock Technical report, Swedish Institute of Computer Science, Box 1263,
  164 29 Kista, Sweden, 2004.

\bibitem{choco}
François Laburthe.
\newblock {CHOCO}: implementing a {CP} kernel.
\newblock In Beldiceanu et~al. \cite{TRICS:00}, pages 71--85.

\bibitem{interval-narrowing}
Olivier Lhomme, Arnaud Gotlieb, and Michel Rueher.
\newblock Dynamic optimization of interval narrowing algorithms.
\newblock {\em Journal of Logic Programming}, 37(1--3):165--183, 1998.

\bibitem{mackworth}
Alan~K. Mackworth.
\newblock Consistency in networks of relations.
\newblock {\em Artificial Intelligence}, 8(1):99--118, 1977.

\bibitem{book}
Kim Marriott and Peter~J. Stuckey.
\newblock {\em Programming with Constraints: an Introduction}.
\newblock The MIT Press, Cambridge, MA, USA, 1998.

\bibitem{ac4}
Roger Mohr and Thomas~C. Henderson.
\newblock Arc and path consistency revisited.
\newblock {\em Artificial Intelligence}, 28:225--233, 1986.

\bibitem{gac4}
Roger Mohr and Gérald Masini.
\newblock Good old discrete relaxation.
\newblock In Yves Kodratoff, editor, {\em Proceedings of the 8th European
  Conference on Artificial Intelligence (ECAI 88)}, pages 651--656, Munich,
  Germany, 1988. Pitmann Publishing.

\bibitem{chaff}
Matthew~W. Moskewicz, Conor~F. Madigan, Ying Zhao, Lintao Zhang, and Sharad
  Malik.
\newblock {Chaff}: Engineering an efficient {SAT} solver.
\newblock In {\em Proceedings of the 38th Design Automation Conference, DAC
  2001}, pages 530--535, Las Vegas, NV, USA, 2001. ACM.

\bibitem{mozart}
{Mozart Consortium}.
\newblock The {Mozart} programming system, 1999.
\newblock Available from \texttt{www.mozart-oz.org}.

\bibitem{regular}
Gilles Pesant.
\newblock A regular language membership constraint for finite sequences of
  variables.
\newblock In Mark Wallace, editor, {\em Tenth International Conference on
  Principles and Practice of Constraint Programming}, volume 3258 of {\em
  LNCS}, pages 482--495. Springer-Verlag, Toronto, Canada, September 2004.

\bibitem{puget-alldifferent}
Jean-Francois. Puget.
\newblock A fast algorithm for the bound consistency of alldiff constraints.
\newblock In {\em Proceedings of the 15th National Conference on Artificial
  Intelligence (AAAI-98)}, pages 359--366, Madison, WI, USA, July 1998. AAAI
  Press/The MIT Press.

\bibitem{Regin94}
Jean-Charles Régin.
\newblock A filtering algorithm for constraints of difference in {CSP}s.
\newblock In {\em Proceedings of the Twelfth National Conference on Artificial
  Intelligence}, volume~1, pages 362--367, Seattle, WA, USA, 1994. AAAI Press.

\bibitem{typer}
Pierre Sav\'{e}ant.
\newblock Constraint reduction at the type level.
\newblock In Beldiceanu et~al. \cite{TRICS:00}, pages 16--29.

\bibitem{SchulteStuckey:TOPLAS:2005}
Christian Schulte and Peter~J. Stuckey.
\newblock When do bounds and domain propagation lead to the same search space?
\newblock {\em Transactions on Programming Languages and Systems},
  27(3):388--425, May 2005.

\bibitem{OPL:1999}
Pascal Van~Hentenryck.
\newblock {\em The {OPL} Optimization Programming Language}.
\newblock The MIT Press, Cambridge, MA, USA, 1999.

\bibitem{ccfd}
Pascal {Van Hentenryck}, Vijay Saraswat, and Yves Deville.
\newblock Constraint processing in cc({FD}).
\newblock Draft, 1991.

\bibitem{ccfd2}
Pascal {Van Hentenryck}, Vijay Saraswat, and Yves Deville.
\newblock Design, implementation and evaluation of the constraint language
  cc({FD}).
\newblock {\em Journal of Logic Programming}, 37(1--3):139--164, 1998.

\bibitem{exactly}
Pascal Van~Hentenryck, Helmut Simonis, and Mehmet Dincbas.
\newblock Constraint satisfaction using constraint logic programming.
\newblock {\em Artificial Intelligence}, 58:113--159, 1992.

\bibitem{eclipse}
Mark Wallace, Stefano Novello, and Joachim Schimpf.
\newblock Eclipse: A platform for constraint logic programming.
\newblock Technical report, IC-Parc, Imperial College, London, GB, August 1997.

\bibitem{Zhou:TPLP:05}
Neng-Fa Zhou.
\newblock Programming finite-domain constraint propagators in action rules.
\newblock {\em Theory and Practice of Logic Programming}, 6(5):483--508,
  September 2006.

\end{thebibliography}

\end{document}